\documentclass{article}

\usepackage{microtype}
\usepackage{graphicx}
\usepackage{booktabs} %

\usepackage{hyperref}

\usepackage[accepted]{icml2023}

\usepackage{amsmath}
\usepackage{amssymb}
\usepackage{mathtools}
\usepackage{amsthm}

\usepackage[capitalize,noabbrev]{cleveref}

\theoremstyle{plain}
\newtheorem{theorem}{Theorem}[section]

\theoremstyle{definition}

\theoremstyle{remark}

\usepackage[textsize=tiny]{todonotes}

\usepackage[T1]{fontenc}
\usepackage[utf8]{inputenc}

\usepackage{here}

\usepackage{caption}
\usepackage{subcaption}

\usepackage{mathtools}
\usepackage{ctable}
\usepackage{footnote}
\usepackage{footnote}
\usepackage{threeparttable}
\usepackage{colortbl}
\usepackage{environ}
\usepackage{booktabs}
\usepackage{diagbox} 
\usepackage{wrapfig} 
\usepackage{bbding}
\usepackage{pifont}
\usepackage{wasysym}
\usepackage{amssymb}
\usepackage{times}
\usepackage{soul}
\usepackage{url}
\usepackage{caption}
\usepackage{graphicx}
\usepackage{amsmath}
\usepackage{amsthm}
\usepackage{cleveref}
\usepackage{tabularx}
\usepackage{multirow}
\usepackage{fancybox}
\usepackage{mathtools}
\usepackage{bold-extra}
\usepackage{lipsum}
\usepackage{siunitx}
\usepackage[export]{adjustbox}
\usepackage{comment}

\DeclarePairedDelimiterX{\infdivx}[2]{(}{)}{%
  #1\;\delimsize\|\;#2%
}

\makeatletter
\let\MYcaption\@makecaption
\makeatother
\usepackage{subcaption}
\makeatletter
\let\@makecaption\MYcaption
\makeatother

\makeatletter
\newcommand{\printfnsymbol}[1]{%
  \textsuperscript{\@fnsymbol{#1}}%
}
\makeatother
\usepackage{stmaryrd}
\usepackage{trimclip}
\makeatletter
\DeclareRobustCommand{\shortto}{%
  \mathrel{\mathpalette\short@to\relax}%
}

\usepackage{etoolbox}

\makeatletter

\newcommand*{\red}[1]{{\textcolor{red}{#1}}}
\newcommand*{\orange}[1]{{\textcolor{orange}{#1}}}
\newcommand*{\blue}[1]{{\textcolor{blue}{#1}}}
\newcommand*{\coolFont}[1]{{\bfseries\scshape{#1}}}

\Crefformat{figure*}{#2Figure~#1#3}
\Crefmultiformat{figure*}{Figures~#2#1#3}{ and~#2#1#3}{, #2#1#3}{ and~#2#1#3}

\crefname{equation}{}{}

\NewEnviron{alignSmall}{%
    \begin{align}
    \BODY
    \end{align}
}
\NewEnviron{gatherSmall}{%
    \begin{gather}
    \BODY
    \end{gather}
}

\usepackage{bussproofs}
\usepackage{stackengine}
\usepackage{varwidth}

\newenvironment{mathprooftree}
  {\varwidth{.9\textwidth}\centering\leavevmode}
  {\DisplayProof\endvarwidth}

\newcommand*{\corpusRT }[0]{RT }
\newcommand*{\corpusRTNL }[0]{RT.PR }

\newcommand*{\corpusFLDAA }[0]{sFLD-crit }
\newcommand*{\corpusFLDRTDpthsRT }[0]{sFLD-impl }

\newcommand*{\corpusSFLD }[0]{\coolFont{sFLD} }
\newcommand*{\corpusFLDRTZero }[0]{FLD-impl.0 }

\newcommand*{\corpusFLDTwo }[0]{FLD.2 }
\newcommand*{\corpusFLDThree }[0]{FLD.3 }
\newcommand*{\corpusFLDFour }[0]{FLD.4 }

\newcommand*{\corpusFLDDFive }[0]{\coolFont{FLD.D5} }

\newcommand*{\corpusFLDRTZeroWOS}[0]{FLD-impl.0}
\newcommand*{\corpusFLDRTOneWOS}[0]{FLD-impl.1}
\newcommand*{\corpusFLDZeroWOS}[0]{FLD.0}
\newcommand*{\corpusFLDOneWOS}[0]{FLD.1}
\newcommand*{\corpusFLDTwoWOS}[0]{FLD.2}
\newcommand*{\corpusFLDThreeWOS}[0]{FLD.3}
\newcommand*{\corpusFLDFourWOS}[0]{FLD.4}

\newcommand*{\corpusRTDFiveWOS}[0]{RT.D5}
\newcommand*{\corpusFLDDFiveWOS}[0]{\coolFont{FLD.D5}}

\newcommand*{\captionFewShot}[1]{Few-shot proof accuracies of provers transferred among corpora that differ in #1.}
\newcommand*{\captionFewShotSameDepthDistrib}[0]{Few-shot proof accuracies of provers transferred among \coolFont{sFLD} and baseline corpora. For fair comparison, all the corpora have the same depth distribution (except \corpusFLDAA that cannot form multistep easily, see \cref{appendix:sec:corpora})}
\newcommand*{\captionFewShotAnswerAccuracy}[1]{Few-shot answer accuracies of provers transferred among corpora that differ in #1.}

\icmltitlerunning{Learning Deductive Reasoning from Synthetic Corpus based on Formal Logic}

\begin{document}

\twocolumn[
\icmltitle{Learning Deductive Reasoning from Synthetic Corpus based on Formal Logic}

\begin{icmlauthorlist}
\icmlauthor{Terufumi Morishita}{comp}
\icmlauthor{Gaku Morio}{comp}
\icmlauthor{Atsuki Yamaguchi}{comp}
\icmlauthor{Yasuhiro Sogawa}{comp}
\end{icmlauthorlist}

\icmlaffiliation{comp}{Hitachi, Ltd. Research and Development Group, Kokubunji, Tokyo, Japan}

\icmlcorrespondingauthor{Terufumi Morishita}{terufumi.morishita.wp@hitachi.com}

\icmlkeywords{Machine Learning, ICML}

\vskip 0.3in
]

\printAffiliationsAndNotice{}  %

\begin{abstract}
We study a synthetic corpus based approach for language models (LMs) to acquire logical deductive reasoning ability.
The previous studies generated deduction examples using specific sets of deduction rules.
However, these rules were limited or otherwise arbitrary, limiting the generalizability of acquired reasoning ability.
We rethink this and adopt a well-grounded set of deduction rules based on formal logic theory, which can derive any other deduction rules when combined in a multistep way.
Then, using the proposed corpora, which we name \coolFont{FLD} (\coolFont{F}ormal \coolFont{L}ogic \coolFont{D}eduction), we first evaluate and analyze the logical reasoning ability of the latest LLMs.
Even GPT-4 can solve only half of the problems, suggesting that pure logical reasoning isolated from knowledge is still challenging for the LLMs, and additional training specialized in logical reasoning is indeed essential.
We next empirically verify that LMs trained on \coolFont{FLD} corpora acquire more generalizable reasoning ability.
Furthermore, we identify the aspects of reasoning ability on which deduction corpora can enhance LMs and those on which they cannot, and discuss future directions on each aspect.
The released corpora serve both as learning resources and as challenging benchmarks.
\end{abstract}

\section{Introduction}  \label{sec:intro}

\begin{figure*}[t!]
    \centering
    \includegraphics[width=1.0\linewidth]{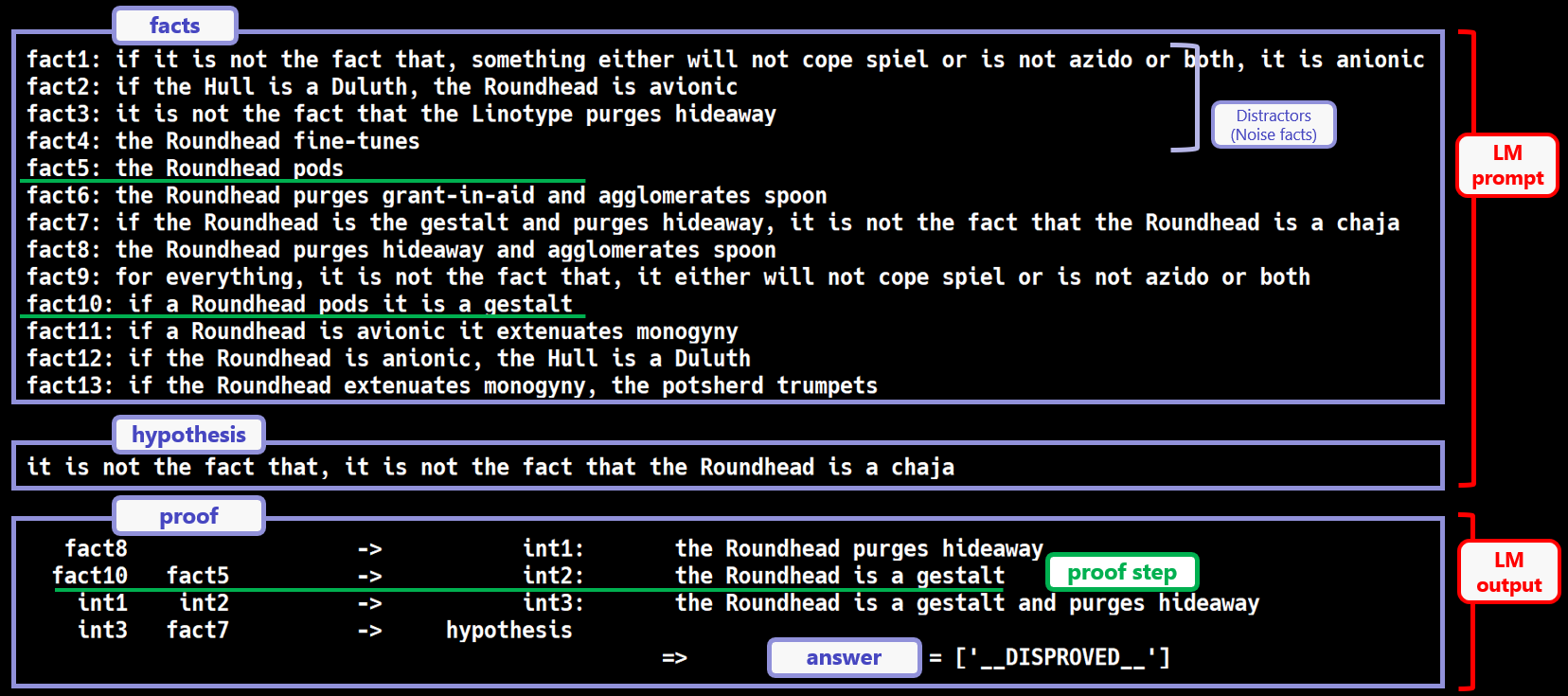}
    \vspace{-7mm}
    \caption{
        A deduction example generated by \coolFont{FLD}.
        Given a set of facts and a hypothesis, an LM is required to generate proof steps to (dis-)prove the hypothesis and an answer.
        The proof is constructed from a theoretically well-grounded set of deduction rules (i.e., the axioms of first-order predicate logic).
        Note that the facts are randomly constructed except logical structures (i.e., they have no semantics) so that referring to existing knowledge never helps solve the task.
        \label{fig:FLD_example}
    }
    \vspace{-5mm}
\end{figure*}

Building a machine that reasons with abundant knowledge has been the Holy Grail since the early era of artificial intelligence \cite{Mccarthy1959ProgramsWC}.
Recent language models (LMs) have taken a step toward this goal, as they demonstrated extensive factual and commonsense knowledge obtained through large-scale pre-training.
Even so, LMs still struggle with logical reasoning \cite{askell2020gpt,rae2021scaling,razeghi2022impact,liu2023evaluating,turpin2023language,lanham2023measuring,wu2023reasoning,hodel2023response,dziri2023faith,dasgupta2023language}, as it demands rigid manipulation of logical rules rather than merely referring to the existing knowledge.

LMs have acquired their knowledge from a lot of high-quality examples in human-written texts \cite{devlin2019bert}.
Conversely, their poor logical reasoning ability suggests the lack of high-quality examples of logical reasoning in the texts.
This is not a surprise given that humans usually think reflexively rather than logically step by step \cite{kahneman2011thinking}.
The consideration here suggests a straightforward strategy to endow LMs with logical reasoning ability: create corpora that include many examples of valid logical reasoning, and train LMs on them.

One such corpus is the recently proposed RuleTaker \cite{clark2020transformers}.
RuleTaker is composed of synthetically generated examples of multistep deductive reasoning.%
A deduction example requires an LM to generate logical steps to (dis-)prove a given hypothesis based on a given set of facts.
Each logical step must follow deduction rules of the implication kind, such as  $\forall x F(x) \to	G(x), F(a) \vdash G(a)$ (here, $\vdash$ means ``derives'').
The facts are randomly constructed except the logical structure (i.e., they have no semantics), therefore referring to existing knowledge never helps solve the task.
Artificial Argument Corpus (AACorpus) \cite{betz-etal-2021-critical} is another synthetic corpus composed of single-step deduction examples.
AACorpus adopted a set of hand-selected deduction rules useful for critical thinking, such as contraposition $\mathcal{F} \to \mathcal{G} \vdash \neg \mathcal{G} \to \neg \mathcal{F}$ ($\neg$ is negation).
All these corpora teach deductive reasoning, one of the most universally used logical reasoning.

However, it is still an open question whether this research direction will genuinely lead to the improvement of deductive reasoning ability.
First, the deduction rules used in the previous corpora were limited or otherwise arbitrary.
This can limit the generalizability of the acquired deductive reasoning ability since complex real-world reasoning can require various deduction rules.
Second, it has not yet been studied what aspect of deductive reasoning ability deduction corpora can enhance LMs.
Such aspects include the ability to solve many-step reasoning, and an understanding of diverse linguistic expressions of logical statements.
This investigation is essential to discuss the future directions.

This paper aims to answer these questions.
First, we rethink the choice of deduction rules by leveraging the formal logic theory (\cref{sec:formal_logic}).
According to formal logic, there are infinite valid deduction rules, including but not limited to the ones used in the previous corpora.
However, among them, there is a set of atomic deduction rules called \textit{the axioms}, and any other valid deduction rules can be derived by multistep deductions constructed from the axioms (\textit{completeness}).
Therefore, \textit{the axioms are the most generalizable to various deduction rules}.
As the sets of deduction rules used in the previous corpora lack this property, we propose a new framework named \coolFont{FLD} (\coolFont{F}ormal \coolFont{L}ogic \coolFont{D}eduction), which generates deduction examples constructed from the axioms.

Then, we first investigate how well current (L)LMs perform logical reasoning (\Cref{sec:how_well}).
We find that even the most powerful LLM, GPT-4 \cite{OpenAI2023GPT4TR}, can solve only half of the problems. 
More fine-grained analyses reveal several phenomena, notably that the LLMs do not reason faithfully following the ``reasoning steps'' they themselves generate.
Overall, the results here suggest that pure logical reasoning isolated from knowledge is still challenging for latest LLMs, and additional training specialized in logical reasoning should be essential.

Next, we show that the training on \coolFont{FLD} is effective  (\cref{sec:effectiveness_of_FLD}).
We trained LMs on \coolFont{FLD} and measured their performance on two types of benchmarks:
one is deduction corpora themselves and the other is human-authored EntailmentBank (EB) \cite{dalvi-etal-2021-explaining}, which requires more complex real-world reasoning.
The resulst are promising as the LMs outperformed baselines on both benchmarks.

Finally, we identify the aspects of deductive reasoning ability on which deduction corpora are beneficial (\cref{sec:synthetic_transfer}).
We analyzed each aspect separately by using a comprehensive set of ``ablation corpora'', where one corpus emphasizes a specific aspect different from those emphasized by the other corpora.
The results suggest that deduction corpora are beneficial in many aspects, but they alone are not enough for some aspects.
Finally, on the basis of the results, we discuss the future directions for each aspect (\cref{sec:future_work}).

We summarize our contributions as follows:
\vspace{-3mm}
\begin{itemize}
\vspace{-0.5\baselineskip} %
\setlength{\itemsep}{-1mm}
\setlength{\leftskip}{-1mm}
    \item We propose\footnote{\scriptsize Available at: \texttt{\url{https://github.com/hitachi-nlp/FLD}} \label{note1} } a deduction corpus generation framework \coolFont{FLD} (\cref{sec:FLD}). \coolFont{FLD} is the first to leverage formal logic theory, adopting a well-grounded set of deduction rules that generalizes the best to other deduction rules.
    \item We evaluate and analyze the logical reasoning ability of latest LLMs (\Cref{sec:how_well}). Even GPT-4 can solve only half of the problems, suggesting that pure logical reasoning isolated from knowledge is still challenging for LLMs.
    \item We empirically verify that LMs trained on \coolFont{FLD} corpora acquire more generalizable deductive reasoning ability than the baselines without such training (\cref{sec:effectiveness_of_FLD}).
    \item We analyze each aspect of deductive reasoning and provide the future directions for applying deduction corpora or other approaches for them (\cref{sec:synthetic_transfer,sec:future_work}).
    \item We release\footref{note1} the \coolFont{FLD} corpora, which serve both as learning resources and as challenging benchmarks.
\end{itemize}

\section{Preliminaries: Formal Logic}   \label{sec:formal_logic}

Let us consider the following single-step deductive reasoning:
\vspace{-2mm}
\begin{equation}
    \begin{mathprooftree}
        \AxiomC{\stackanchor{\small The Earth revolves}{\small around the sun}}
        \AxiomC{\stackanchor{\small If the Earth revolves around the sun,}{\small the Earth has seasons.}}
        \BinaryInfC{\small The Earth has seasons.}
    \end{mathprooftree}
    \label{eq:argument:first}
\end{equation}
This deduction step derives the conclusion, written under the bar, from the two premises.
Next, consider another step:
\vspace{-2mm}
\begin{equation}
    \begin{mathprooftree}
        \AxiomC{\stackanchor{\small The Earth revolves}{\small around the sun}}
        \AxiomC{\stackanchor{\small If the Earth revolves around the sun,}{\small the Earth does not have seasons.}}
    \BinaryInfC{\small The Earth does not have seasons.}
    \end{mathprooftree}
    \label{eq:argument:second}
\end{equation}
In this step, one of the premises (i.e., ``If the Earth revolves around the sun, the Earth does not have seasons'') is false.
However, \textit{if the premise had been true}, we can still derive the conclusion.
Thus, in formal logic, this step is still valid the same as \Cref{eq:argument:first}.
We can abstract \cref{eq:argument:first,eq:argument:second} into a deduction rule as:
\vspace{-2mm}
\begin{equation}
    \begin{mathprooftree}
        \AxiomC{$\mathcal{F}$}
        \AxiomC{$\mathcal{F} \rightarrow \mathcal{G}$}
        \RightLabel{\stackanchor{modus ponens}{}}
        \BinaryInfC{$\mathcal{G}$}
    \end{mathprooftree}
    \label{eq:argument:modus_ponens}
\end{equation}
The deduction rule of this form is called \textit{modus ponens}.

While modus ponens is the most intuitive deduction rule, many others exist.
For example, a famous syllogism is:
\vspace{-2mm}
\begin{equation}
    \begin{mathprooftree}
    \Axiom $(\mathcal{F} \rightarrow \mathcal{G}) \fCenter \land (\mathcal{G} \rightarrow \mathcal{H})$
    \RightLabel{\stackanchor{syllogism}{}}
    \UnaryInf $\mathcal{F} \fCenter \rightarrow \mathcal{H}$
    \end{mathprooftree}
    \label{eq:argument:syllogism}
\end{equation}
\vspace{-2mm}
The other example below defines the meaning of $\land$ formally:
\vspace{-1mm}
\begin{equation}
    \begin{mathprooftree}
    \Axiom $(\mathcal{F} \fCenter \land \mathcal{G})$
    \UnaryInf $\fCenter \mathcal{F}$
    \end{mathprooftree}
    \ 
    \begin{mathprooftree}
    \Axiom $(\mathcal{F} \fCenter \land \mathcal{G})$
    \RightLabel{\stackanchor{$\land$-elimination}{}}
    \UnaryInf $\fCenter \mathcal{G}$
    \end{mathprooftree}
    \label{eq:argument:and_elim1}
\end{equation}
Of course, we can consider invalid
\footnote{
A deduction step is invalid when for some truth value assignments, the conclusion is false (=0) even if all the premises are true (=1).
See \cref{tb:truth_table_wrong}.
}
rules, in the sense that the conclusion is not logically deducible from the premises, such as:
\vspace{-2mm}
\begin{equation}
    \begin{mathprooftree}
    \AxiomC{$\mathcal{F}$}
    \AxiomC{$(\mathcal{F} \lor \mathcal{G})$}
    \BinaryInfC{$\mathcal{G}$}
    \end{mathprooftree}
    \label{eq:argument:incorrect}
\end{equation}
\vspace{-6mm}

Now, from these examples, we obtain some important points of deductive reasoning.
First, deductive reasoning can be defined as a form of thought in which a conclusion is derived from a set of premises following specific deduction rules, such as the ones in \crefrange{eq:argument:first}{eq:argument:incorrect}.
Such deduction rules are called \textit{arguments} in formal logic theory.
Second, whether a deduction rule is valid or not does not depend on \textit{contents} of symbols but only on the \textit{superficial form} of the symbolic sequence composed of the premises to the conclusion.
For example, as stated above, \cref{eq:argument:modus_ponens} is valid regardless of the actual content of $\mathcal{G}$, such as $\mathcal{G}$=``(\dots), the Earth has seasons.'' in \cref{eq:argument:first} and $\mathcal{G}$=``(\dots), the Earth does not have seasons.'' in \cref{eq:argument:second}.
This enables us to regard all deduction rules simply as symbolic rules such as \crefrange{eq:argument:modus_ponens}{eq:argument:incorrect}.
Third and as one conclusion of the second point, the symbols such as $\mathcal{F}$ and $\mathcal{G}$ can be arbitrary compounds of other formulas such as $\mathcal{F}$=$(A \land B)$ and $\mathcal{F}$=$\forall x, A(x) \rightarrow B(x)$.
Finally, since we can consider infinite patterns of formulas as premises and a conclusion, we have infinite patterns of deduction rules (including both valid and invalid deduction rules).

\begin{figure}[t]
    \centering
    \includegraphics[width=0.93\linewidth]{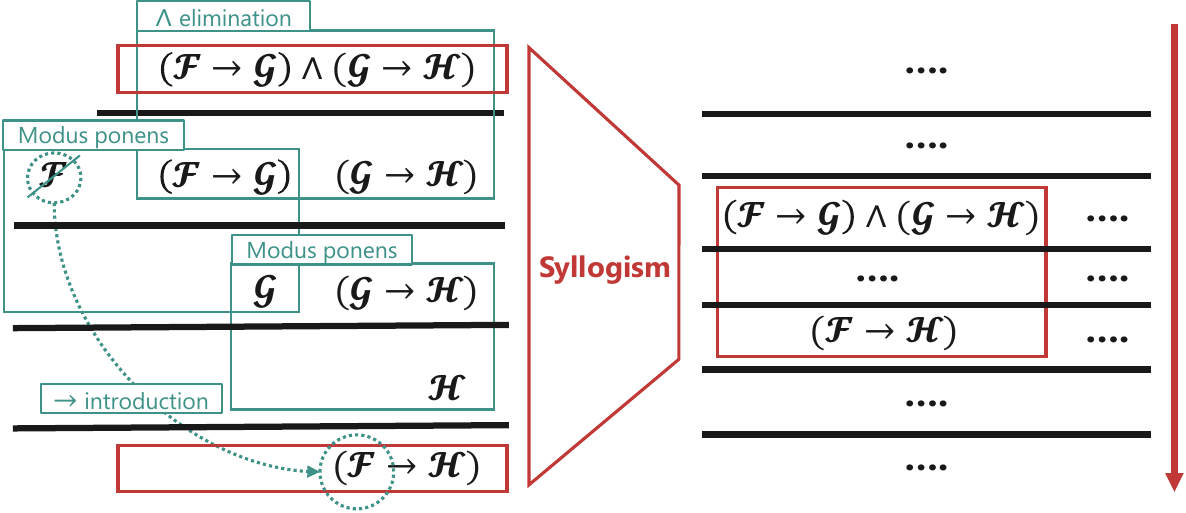}
    \vspace{-2mm}
    \caption{
        An example of multistep deduction constructed from the axioms.
        \textbf{(Left)} shows the derivation of a syllogism.
        \textbf{(Right)} illustrates that deduction with more steps can express deductions that use a syllogism as a given rule.
        \label{fig:proof_syllogism}
    }
    \vspace{-7mm}
\end{figure}

Next, we consider multistep deductions.
\Cref{fig:proof_syllogism} shows that syllogism rule can be derived by the multistep deduction constructed from other ``atomic'' deduction rules.
(For other examples, \cref{appendix:argument:proofs} shows the derivations of the deduction rules used in the previous corpora.)
Indeed, in formal logic, there is a set of atomic deduction rules called \textit{the axioms} (listed in \cref{appendix:argument:axioms}), and the following is known
\footnote{
We limit our focus to first-order predicate logic in this paper.
}
:
\vspace{-1mm}
\begin{theorem}[Completeness of first-order predicate logic \citep{godel1930uber}]
    \label{theorem:completeness}
    Any valid
    \footnote{
    A deduction rule is valid when for all truth value assignments, the conclusion is true (=1) if all the premises are true.
    See \cref{tb:truth_table_syllogism}.
    }
    deduction rule is derivable by multistep deduction constructed from the axioms.
    Furthermore, any deduction rule derivable by multistep deduction constructed from the axioms is valid.
\end{theorem}
\vspace{-3mm}
Here we have come to the core of formal logic: multistep deduction constructed from the axioms.
Thanks to the completeness, all valid deduction rules can be derived in this way.
As a consequence, \textit{multistep deduction constructed from the axioms can express multistep deduction constructed from any other deduction rules}, as illustrated in \cref{fig:proof_syllogism} (right).

\vspace{-3mm}
\section{Generating Formal Logic Deduction Corpus}   \label{sec:FLD}
\vspace{-1mm}

\begin{figure*}[t!]
    \vskip 0pt
    \centering
    \includegraphics[width=1.0\linewidth]{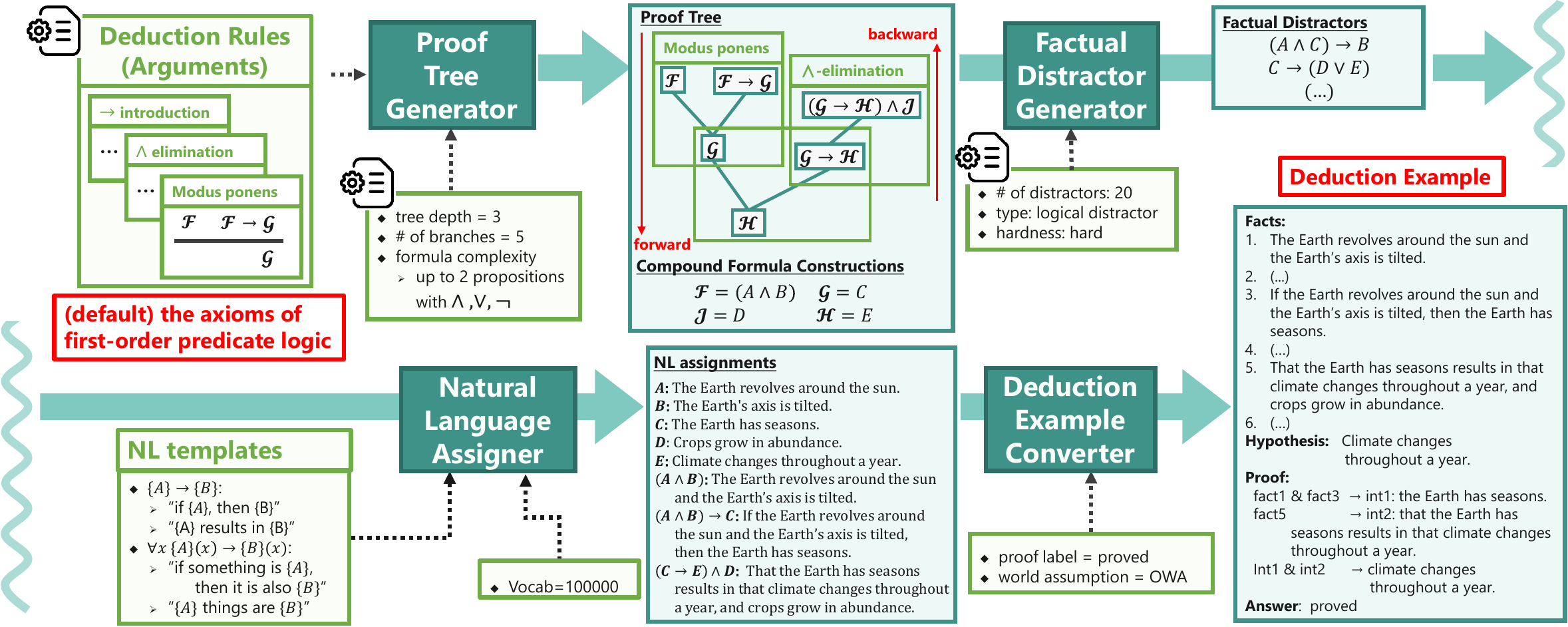}
    \vspace{-3mm}
    \caption{
        An overview of the proposed framework \coolFont{FLD}, which generates logical deduction examples constructed from the axioms of first-order predicate logic.
        \coolFont{FLD} is flexible to generate various patterns of corpora for analysis. 
        Note that the actual natural language assignments are constructed as random as possible to assess rigid logical reasoning ability isolated from knowledge (see \Cref{sec:FLD}).
        \label{fig:framework_overview}
    }
\end{figure*}

\begin{table*}[h!]
    \scriptsize
    \centering
    \tabcolsep 3pt
    \caption{
        A comparison of \coolFont{FLD} with the previous studies.
        \coolFont{FLD} is flexible to generate various patterns of corpora for analysis.
        \red{\checkmark} means controllable and extensible by an external template file.
        \orange{\checkmark} means controllable by an option.
        \label{tb:framework_comparison}
    }

\begin{tabular}{@{}cccccccc@{}}
\toprule
                                                                                  & \begin{tabular}[c]{@{}c@{}}Deduction\\ Rules\end{tabular}                                & \begin{tabular}[c]{@{}c@{}}Proof Tree\\ Depth (upto)\end{tabular}                   & \begin{tabular}[c]{@{}c@{}}Proof Tree\\ Branches\end{tabular}                       & \begin{tabular}[c]{@{}c@{}}Formula\\ Complexity\end{tabular}                              & \begin{tabular}[c]{@{}c@{}}\# of\\ Distractors (up to)\end{tabular}                 & \begin{tabular}[c]{@{}c@{}}Linguistic\\ Diversity\end{tabular}                    & \begin{tabular}[c]{@{}c@{}}Proof\\ Labels\end{tabular}                                  \\ \midrule
\begin{tabular}[c]{@{}c@{}}RuleTaker\\ \cite{clark2020transformers}\end{tabular}  & implication                                                                              & 5                                                                                   & A few                                                                               & complex                                                                                   & $\sim$20                                                                            & \begin{tabular}[c]{@{}c@{}}less (RuleTaker) /\\ more (ParaRules)\end{tabular}     & \begin{tabular}[c]{@{}c@{}}provable /\\ disprovable /\\ unknown\end{tabular}            \\ \midrule
\begin{tabular}[c]{@{}c@{}}AACorpus\\ \cite{betz-etal-2021-critical}\end{tabular} & \begin{tabular}[c]{@{}c@{}}\red{\checkmark}\\ (default = critical thinking)\end{tabular} & 1                                                                                   & 1                                                                                   & \begin{tabular}[c]{@{}c@{}}\orange{\checkmark}\\ (\orange{simple / complex})\end{tabular} & 0                                                                                   & \begin{tabular}[c]{@{}c@{}}\red{\checkmark}\\ (default = less)\end{tabular}       & \begin{tabular}[c]{@{}c@{}}provable /\\ disprovable\end{tabular}                        \\ \midrule
\coolFont{FLD}                                                                    & \begin{tabular}[c]{@{}c@{}}\red{\checkmark}\\ (default = \red{the axioms})\end{tabular}  & \begin{tabular}[c]{@{}c@{}}\orange{\checkmark}\\ (\orange{can be any})\end{tabular} & \begin{tabular}[c]{@{}c@{}}\orange{\checkmark}\\ (\orange{can be any})\end{tabular} & \begin{tabular}[c]{@{}c@{}}\orange{\checkmark}\\ (\orange{simple / complex})\end{tabular} & \begin{tabular}[c]{@{}c@{}}\orange{\checkmark}\\ (\orange{can be any})\end{tabular} & \begin{tabular}[c]{@{}c@{}}\red{\checkmark}\\ (default = \red{more})\end{tabular} & \begin{tabular}[c]{@{}c@{}}\orange{\checkmark}\\ (\orange{can choose any})\end{tabular} \\ \bottomrule
\end{tabular}

\vspace{-5mm}
\end{table*}

The previous deduction corpora \cite{clark2020transformers,betz-etal-2021-critical} used limited or arbitrary sets of deduction rules.
However, as we saw in \cref{sec:formal_logic}, the axioms should be the most generalizable to various deduction rules.
Thus, we propose a framework named \coolFont{FLD} (\coolFont{F}ormal \coolFont{L}ogic \coolFont{D}eduction), which generates examples of multistep deduction constructed from the axioms.

Another important feature of \coolFont{FLD} is that the statements in the examples are constructed randomly except logical structures (i.e., they have no semantics), so that referring to existing knowledge never helps solve the task, but only adhering to deduction rules solves the task.
Finally, we designed \coolFont{FLD} to be highly flexible as in \cref{tb:framework_comparison}, so that we can generate and analyze various patterns of corpora.

We show generated examples in \cref{fig:FLD_example,appendix:fig:FLD_output_sample}.
Below, we overview each module.
For intuitive understanding, refer to the corresponding part of \cref{fig:framework_overview}.
For the detailed implementations, refer to \cref{appendix:sec:FLD}.

\subsection{Proof Tree Generation via Random Forward-/Backward- Deduction}   \label{sec:FLD_proof_tree_generation}

RuleTaker \cite{clark2020transformers} generates deductive proof trees by first randomly generating various formulas and second running a logical solver library on them to find occasionally emerged deductive relationships among them.
However, since we rely on an external solver, we cannot specify the set of deduction rules used in proof trees (and thus we cannot specify the axioms, especially.).
Further, since we rely on the randomness, we cannot control the complexity of a proof tree, i.e., the depth and the number of leaves.

Thus, we decided to take another approach.
We invented a module (``Proof Tree Generator'' in \cref{fig:framework_overview}) that generates a proof tree through a random deduction process by using a set of deduction rules specified by a user.
A user can specify the deduction rules in a template rule file, as exemplified in \cref{appendix:fig:argument_template}.
At each forward- or backward- deduction step, the module randomly chooses one deduction rule and joints it to the current proof tree (``forward'' and ``backward'' in the figure).
The numbers of forward- and backward- steps control the tree's depth and number of leaves, respectively.

Once the structure of the proof tree is constructed, we construct the compound formulas at the tree nodes, such as $\mathcal{F}, \mathcal{G}$.
Since these formulas are arbitrary (\cref{sec:formal_logic}), we randomly combine atomic formulas such as $A$ and $B$ using logical operators $\land, \lor, \neg$.
To avoid over complications, we limit the number of atomic formulas in each compound formula up to three.
The resulting formulas are like $\mathcal{F}=(\neg A\land B)$.

\subsection{Factual Distractor Generation}   \label{sec:FLD_distractor_generation}

In a realistic scenario of logical reasoning, since the facts are collected by possibly incomplete retrieval systems rather than given, LMs have to correctly choose only the relevant facts under the existence of many irrelevant facts.
To imitate this scenario, we add distractor facts to each deduction example (``Factual Distractor Generator'' in \cref{fig:framework_overview}).
The distractor facts are formulas that are similar to the gold facts in their logical form.
For example, for the gold fact $(A \land B) \rightarrow C$, formulas such as $(A \land C) \rightarrow B$ can be distractors.
We also implemented several other types of distractors and use the mixture of them.

\subsection{Natural Language Assignment}   \label{sec:FLD_NL_assigner}
\normalsize
We assign one natural language sequence to each formula of tree nodes and of distractors (``Natural Language Assigner'' in \cref{fig:framework_overview}).
Inspired by \citet{betz-etal-2021-critical}, we take a template based approach.
For each formula, we prepare several templates via an external template file (exemplified in \cref{appendix:fig:translation_template}) such as follows:

\vspace{-5mm}
{\small
  \setlength{\abovedisplayskip}{2pt}
  \setlength{\belowdisplayskip}{\abovedisplayskip}
  \setlength{\abovedisplayshortskip}{0pt}
  \setlength{\belowdisplayshortskip}{3pt}
  \begin{align}
        A \rightarrow B: & \ \text{``If A, then B.''}, \ \text{``A leads to B.''}\nonumber \\
        F(a) \rightarrow G(b):  & \ \text{``If a F, then b G.''}, \ \text{``When a F, b G.''} \nonumber
  \end{align}
}%
Then, we randomly choose one from them.
Note that since the templates can be nested, the number of resulting patterns are combinatorially diverse.

Next, we assign natural language statements to atomic components such as $A, B, F, G, a, b$.
Here, we come back to the important point in deductive reasoning discussed in \cref{sec:formal_logic}: that the validity of deduction does not depend on contents of formulas, or in other words, the same deduction can be conducted on the same formulas regardless of their contents.
To reflect this point, we assign a \textit{random} statement constructed (under a certain grammatical constraint) from a full vocabulary to each atomic component; for example:

\vspace{-5mm}
{\small
  \setlength{\abovedisplayskip}{2pt}
  \setlength{\belowdisplayskip}{\abovedisplayskip}
  \setlength{\abovedisplayshortskip}{0pt}
  \setlength{\belowdisplayshortskip}{3pt}
  \begin{align}
    A:\ \text{``an Earthquake occurs''}  &  \ \ \ B:\ \text{``the year ends''} \nonumber \\
    F:\ \text{``run''}    \ \ \ G:\ \text{``answer''} \ \ \   a:\ & \text{``the hamburger''} \ \ \  b:\ \text{``Peter''}  \nonumber
  \end{align}
}%
These randomly constructed statements ensure that referring to existing knowledge never helps solve the task, but only adhering to deduction rules solves the task.

\subsection{Deduction Example Conversion}   \label{sec:FNLN_text_serializer}

We finally make a deductive reasoning example from the outputs of the previous modules (``Deduction Example Converter'' in \cref{fig:framework_overview}).
A deduction example is composed of a set of facts, a hypothesis, a proof sequence, and an answer (``proved'', ``disproved'', or ``unknown'').
This module can make an example of any answer label as follows.
For answer label ``proved'',
(i) we use the root node as the hypothesis,
(ii) we use the leaf nodes of the proof tree and the distractors as the fact set, and
(iii) we use the internal nodes of the proof tree as the proof sequence.
For answer label ``disproved'', we use the negated statement of the root node as the hypothesis so that the hypothesis is disproved by the proof sequence.
For answer label ``unknown'', we randomly drop some of the leaf nodes so that the hypothesis cannot be proved or disproved by the proof sequence.

\section{Experiments}    \label{sec:experiments}

\begin{table*}[t!]
    \centering
    \footnotesize
    \tabcolsep 4.5pt
    \caption{
        The corpora examined in this paper.
        For RuleTaker (``RT''), we used the OWA version introduced by \citet{tafjord-etal-2021-proofwriter}.
        To align conditions as closely as possible across the corpora being compared, we
        (i) generated multiple FLD corpora using the options and template files and
        (ii) added several preprocessings to RuleTaker.
        See \Cref{appendix:sec:corpora} for details.
        \label{tb:corpora}
    }
    \vspace{-3mm}

\begin{tabular}{@{}llllllcl@{}}
\toprule
name                          & \begin{tabular}[c]{@{}l@{}}deduction\\ rules\end{tabular} & \begin{tabular}[c]{@{}l@{}}distractors\\ (up to)\end{tabular} & \begin{tabular}[c]{@{}l@{}}linguistic\\ diversity\end{tabular} & \begin{tabular}[c]{@{}l@{}}formula\\ complexity\end{tabular} & tree depth & \multicolumn{1}{l}{\begin{tabular}[c]{@{}l@{}}tree depth\\ distribution\end{tabular}}              & \begin{tabular}[c]{@{}l@{}}\#  train\\ examples\end{tabular} \\ \midrule
RT (``D0-D3'')                & implication                                                           & $\sim 20$                                                     & less                                                           & complex                                                      & 1--3       & \multirow{6}{*}{\begin{tabular}[c]{@{}c@{}}skewed\\ (biased toward\\   lower depths)\end{tabular}} & 30k                                                          \\
RT.PR (``ParaRules'')         & implication                                                           & $\sim 20$                                                     & more                                                           & complex                                                      & 1--5       &                                                                                                    & 30k                                                          \\
RT.BE (``Birds-Electricity'') & implication                                                           & $\sim 20$                                                     & more                                                           & complex                                                      & 1--3       &                                                                                                    & - (test only)                                                \\ \cmidrule(r){1-6} \cmidrule(l){8-8} 
sFLD-impl                     & implication                                                           & $\sim 20$                                                     & less                                                           & complex                                                      & 1--3       &                                                                                                    & 30k                                                          \\
sFLD-crit                     & critical thinking                                                     & $\sim 20$                                                     & less                                                           & complex                                                      & 1--1       &                                                                                                    & 30k                                                          \\
sFLD-axiom (\coolFont{sFLD})  & the axioms                                                            & $\sim 20$                                                     & less                                                           & complex                                                      & 1--3       &                                                                                                    & 30k                                                          \\ \toprule
RT.D5 (``D0-D5'')             & implication                                                           & $\sim 20$                                                     & less                                                           & complex                                                      & 1--5       & \multirow{9}{*}{uniform}                                                                           & 30k                                                          \\
\coolFont{FLD.D5}             & the axioms                                                            & $\sim 20$                                                     & less                                                           & complex                                                      & 1--5       &                                                                                                    & 30k                                                          \\ \cmidrule(r){1-6} \cmidrule(l){8-8} 
FLD-impl.0                    & implication                                                           & $\sim 20$                                                     & less                                                           & complex                                                      & 1--3       &                                                                                                    & 30k                                                          \\
FLD-impl.1                    & implication                                                           & $\sim 20$                                                     & less                                                           & complex                                                      & 1--8       &                                                                                                    & 30k                                                          \\ \cmidrule(r){1-6} \cmidrule(l){8-8} 
FLD.0                         & the axioms                                                            & 0                                                             & less                                                           & complex                                                      & 1--3       &                                                                                                    & 30k                                                          \\
FLD.1                         & the axioms                                                            & $\sim 20$                                                     & less                                                           & simple                                                       & 1--3       &                                                                                                    & 30k                                                          \\
FLD.2                         & the axioms                                                            & $\sim 20$                                                     & less                                                           & complex                                                      & 1--3       &                                                                                                    & 30k                                                          \\
FLD.3  (\coolFont{FLD})       & the axioms                                                            & $\sim 20$                                                     & more                                                           & complex                                                      & 1--3       &                                                                                                    & 30k                                                          \\
FLD.4 (\coolFont{FLD}$\star$) & the axioms                                                            & $\sim 20$                                                     & more                                                           & complex                                                      & 1--8       &                                                                                                    & 30k                                                          \\ \bottomrule
\end{tabular}

\vspace{-5mm}
\end{table*}

We conducted experiments to verify the effectiveness of \coolFont{FLD}, and to identify the aspects of deductive reasoning ability on which deduction corpora can enhance LMs.
To this end, we examined various deduction corpora shown in \cref{tb:corpora}.
We trained LMs on the deduction corpora and measured their performance on relevant benchmarks.
For reference, we also measured the performance of a LM (T5) without training on the deduction corpora.
We used two types of benchmarks: deduction corpora themselves and human-authored EntailmentBank \cite{dalvi-etal-2021-explaining}.
We briefly explain the setup.
See \cref{appendix:sec:experiments} for the details.

\subsection{Model}   \label{sec:prover}

All the experiments involve generating a proof sequence to (dis-)prove a given hypothesis from a given set of facts.
To tackle the task of this type, we adopt the stepwise prover model from \citet{yang2022nlproofs}.
This prover is a generative model based on T5 \cite{raffel2020exploring}, which generates one proof step at a time.
A proof step represents the chosen premises and the derived (generated) conclusion, such as ``fact1 \& fact3 -> The Earth has seasons''.
The prover continues the generation until the given hypothesis is (dis-)proved.
Finally, the prover outputs an answer labeled as "proved," "disproved," or "unknown."

We also evaluated large language models in a few-shot in-context learning setting.
Specifically, we evaluated GPT-4, GPT-3.5-Turbo \cite{OpenAI2023GPT4TR}, and LongAlpaca-13B \cite{chen2023longlora}.
Each in-context example is a pair of a prompt and an output, as illustrated in \Cref{fig:FLD_example}.
We also added a chain-of-thought \cite{kojima-2022} instruction as ``Show me a step-by-step thought to the hypothesis based on the given set of facts.''
We ran experiments for 10 different sets of in-context examples.

\subsection{Few-shot Transfer to Synthetic Deduction Corpora}   \label{sec:experiments_synthetic_transfer}

The first benchmark is the deduction corpora, which measures rigid logical reasoning ability.
We trained prover LM on a corpus and measured its performance on another corpus.
If LMs have acquired robust deductive reasoning ability, they should transfer well with a small number of examples.
To see this, we used few-shot setting
\footnote{
Zero-shot is not appropriate for transfer among corpora that differ in the sets of deduction rules used in proofs as follows.
Since a proof step is made by a deduction rule, the nature (granularity) of proof steps in one corpus differs much from those in another corpus.
To adjust this artificial difference, LMs need examples of the target corpus.
}.

We measure the performance of the prover on the test split of the target corpus using two types of metrics, answer accuracy and proof accuracy \cite{saha-etal-2020-prover}.
Answer accuracy measures whether predicted answers, each of which is chosen from ``proved'', ``disproved'' and ``unknown'', are correct or not.
Proof accuracy measures whether \textit{both} the predicted answers and the generated proofs are correct or not, offering a stricter and more faithful assessment of reasoning ability.
Our proof accuracy metric is stricter than the original by \citet{saha-etal-2020-prover} (details in our repository\footref{note1}).
We focus our discussion on proof accuracy results, with answer accuracy results presented in \cref{appendix:sec:answer_accuracies}.

We trained prover LM (T5-base) on the training split of each source corpus for 20k steps with a batch size of 64 and learning rate of 1e-4.
Then we fine-tuned the prover LM on $1\%$ subset ($300$ examples) of the training split of the target corpus.

\subsection{Transfer to EntailmentBank}   \label{sec:experiments_NL_transfer}

EntailmentBank (EB) \cite{dalvi-etal-2021-explaining} is a recently proposed challenging benchmark.
The proof trees in the EB dataset are human-authored rather than synthetically generated.
Further, each proof step can be rough entailment instead of a rigid logical step.
Thus, EB measures logical reasoning ability in a more real-world scenario.

We used all the three tasks of EB, which differ in the property of a given fact set: Task1 does not include distractors, Task2 includes distractors, and Task3 includes sentences retrieved from worldTree V2 \cite{xie-etal-2020-worldtree}.

As stated above, the nature of proof steps in EB differs much from the nature of those in deduction corpora.
Thus, it is difficult for prover LMs trained on deduction corpora to transfer to EB with a small number of examples.
Thus, we fine-tuned the provers using all the EB examples.

We trained a prover LM (T5-large) on a source deduction corpus for 10k steps and fine-tuned it on each EB corpus for 10k steps.
For all the training, the batch size was 64 and the learning rate was 5e-5, except EB-task2 where the learning rate of 2.5e-5 was used.
For EB-task3, we used the prover trained on task2, following \citet{dalvi-etal-2021-explaining}.
Given the challengingness of EB, we used the additional RoBERTa \cite{liu-et-al-2019-roberta} based proof step verifier proposed in \citet{yang2022nlproofs}.
We measured the performance of the provers on the test split of EB by the official metric of ``AllCorrect'' proof accuracy \cite{dalvi-etal-2021-explaining}.

\section{How Well do LMs Solve Logic?}    \label{sec:how_well}

\subsection{Performance of LMs in Fine-tuned Setting}

\vspace{-4mm}

\begin{table}[H]
\centering
\scriptsize

\caption{
    Proof accuracy of a prover fully fine-tuned using all the dataset examples on each corpus.
    \red{Note that we released\footref{note1} the updated, version 2 of \coolFont{FLD} corpora, on which the provers perform slightly better (See \Cref{appendix:sec:release} for details).}
    \label{tb:synthetic_transfer_challengingness}
}
\vspace{-2mm}

\begin{tabular}{llll}
\toprule

\multicolumn{2}{c}{RuleTaker} & \multicolumn{2}{c}{FLD} \\
\cmidrule(l{\tabcolsep}r{\tabcolsep}){1-2} \cmidrule(l{\tabcolsep}){3-4}

                       RT &                     RT.PR &                     \coolFont{FLD} &                     \coolFont{FLD}$\star$ \\
\midrule
 \cellcolor{blue!58} 92.4 &  \cellcolor{blue!60} 93.9 &  \cellcolor{blue!35} 66.4 &  \cellcolor{blue!10} 37.7 \\
\bottomrule
\end{tabular}

\end{table}
\vspace{-6mm}

First, we show how well LMs solve logic of each deduction corpus (\cref{tb:synthetic_transfer_challengingness}).
As shown, while the fully fine-tuned provers performed well on RuleTaker, they performed poorer on FLD.
One possible reason is as follows.
First, since a proof tree is constructed from the combination of a deduction rule chosen at each level of the tree, the number of possible proof tree patterns can be estimated (very roughly) as $\mathcal{O}(\mathcal{A}^d)$, where $\mathcal{A}$ is the number of deduction rule choices and $d$ is the proof tree depth.
Next, while RuleTaker uses only a few deduction rules ($\mathcal{A} = 2$) of implication type shown in \cref{appendix:argument:implication}, FLD uses various deduction rules ($\mathcal{A} \sim 10$) of the axioms shown in \cref{appendix:argument:axioms}.
Thus, FLD includes exponentially more diverse patterns of proof trees, which makes FLD more challenging.
Indeed, when we enlarge the maximum tree depth from $d$=3 to $d$=8 (\coolFont{FLD} to \coolFont{FLD}$\star$), the corpus became extremely more challenging due to the exponentially more diverse proof tree patterns.
See \cref{appendix:sec:synthetic_transfer_challengingness} for further detailed analysis.

\subsection{Performance of LLMs in Few-shot Setting}   \label{sec:LLMs}

\begin{table}[H]
\centering
\small

\caption{
    Performances of LLMs in a 10-shot in-context learning setting.
    Chain-of-thought-like instruction was also used.
    For reference, the performances of random guesses and T5 fine-tuned on all the 30,000 examples are also shown.
    \red{We used the version 2 corpora (See \Cref{appendix:sec:release}).}
    \label{tb:LLMs}
}
\vspace{-2mm}

\begin{tabular}{@{}lcclcc@{}}
\toprule
               & \multicolumn{2}{c}{proof accuracy}     &  & \multicolumn{2}{c}{answer accuracy}    \\ \cmidrule(lr){2-3} \cmidrule(l){5-6} 
               & \coolFont{FLD} & \coolFont{FLD$\star$} &  & \coolFont{FLD} & \coolFont{FLD$\star$} \\ \midrule
random guess   & 0.0            & 0.0                   &  & 33.3           & 33.3                  \\
T5(fine-tuned) & 75.8           & 44.4                  &  & 91.6           & 72.2                  \\ \midrule
LongAlpaca-13B & 0.0            & 0.0                   &  & 21.2           & 19.6                  \\
GPT-3.5-Turbo  & 0.0            & 2.0                   &  & 35.8           & 37.6                  \\
GPT-4          & 12.8           & 3.2                   &  & 52.4           & 49.4                  \\ \bottomrule
\end{tabular}

\end{table}

\Cref{tb:LLMs} show the performance of LLMs measured under 10-shot settings.
As seen from the proof accuracy results, even the most powerful LLM, GPT-4, performed very poorly.
Further, even when measured by answer accuracy, which is more lenient than proof accuracy, GPT-4 solved only half of the problems.

The challengingness of \coolFont{FLD} could be attributed to its counterfactual nature.
The facts in the deduction examples are randomly constructed except for logical structure (i.e., they have no semantics).
Due to this nature, an LLM can \textit{never} rely on its pre-acquired knowledge to solve the task, but it has to adhere to the deduction rules.
Such a kind of task should not be covered by pre-training on human-written texts.
The poor performance of LLMs under the counterfactual setting is consistent with the recent observations by \cite{razeghi2022impact,wu2023reasoning,hodel2023response,dasgupta2023language}.

As can be seen, the answer accuracy performance deviated much from the proof accuracy performance.
This gap can be partly explained by the ``random guess factor'', but not entirely.
We analyzed the proofs generated by the LLMs and found that \textit{they sometimes generated a correct answer with an incorrect proof}.
This suggests that the LLMs do not always faithfully follow the ``logical steps'' they themselves generate.
These results align with recent findings by \citet{turpin2023language,lanham2023measuring}.

Finally, we manually analyzed and categorized the common errors of GPT-4, as follows.
The first is \textit{logical hallucination}, where the generated conclusion is not logically deducible from the chosen facts.
Second, GPT-4 sometimes chooses facts from distractors that are irrelevant to the proof.
Taking a closer look, GPT-4 occasionally misinterprets logical operators, especially double-negation $\neg\neg$.
Interestingly, GPT-4 seldom answered ``unknown'', i.e., it answered ``proved'' or ``disproved'' even if the given facts are not sufficient to either prove or disprove the hypothesis.
Overall, these errors suggest that LLMs still miss the fundamentals of logical reasoning.

\vspace{-2mm}
\section{How Effective is Formal Logic Deduction?}    \label{sec:effectiveness_of_FLD}

\vspace{-1mm}

\subsection{Benchmarking by Deduction Corpora}   \label{sec:effectiveness_of_FLD_arguments}

\vspace{-3mm}

\begin{table}[h]
\centering
\scriptsize
\tabcolsep 2.0pt

\caption{
    \captionFewShotSameDepthDistrib
    \label{tb:synthetic_transfer_arguments}
}
\vspace{-2mm}

\begin{tabular}{llllllll}

\toprule

{} & {}  & {} & \multicolumn{5}{c}{Source corpus} \\
\cmidrule(l{\tabcolsep}r{\tabcolsep}){4-8}

                      & {} &                        T5 &                        RT &                     RT.PR &                  sFLD-impl &                  sFLD-crit &               \textbf{sFLD} \\
\midrule

\multirow{6}{*}{\begin{tabular}[c]{@{}l@{}} Target \\ corpus\end{tabular}} & RT          &  \cellcolor{blue!29} 70.1 &  \cellcolor{blue!57} 92.4 &  \cellcolor{blue!56} 91.3 &  \cellcolor{blue!37} 76.2 &  \cellcolor{blue!35} 74.4 &  \cellcolor{blue!38} 76.7 \\
                      & RT.PR       &  \cellcolor{blue!22} 64.3 &  \cellcolor{blue!56} 91.3 &  \cellcolor{blue!59} 93.9 &  \cellcolor{blue!33} 73.4 &  \cellcolor{blue!26} 67.5 &  \cellcolor{blue!33} 72.9 \\
                      & RT.BE       &  \cellcolor{blue!12} 56.1 &  \cellcolor{blue!52} 88.3 &  \cellcolor{blue!52} 88.2 &  \cellcolor{blue!36} 75.2 &  \cellcolor{blue!41} 79.4 &  \cellcolor{blue!48} 85.0 \\
                      & sFLD-impl    &  \cellcolor{blue!14} 58.4 &  \cellcolor{blue!25} 66.7 &  \cellcolor{blue!24} 65.9 &  \cellcolor{blue!45} 82.2 &  \cellcolor{blue!26} 67.3 &  \cellcolor{blue!43} 80.7 \\
                      & sFLD-crit    &  \cellcolor{blue!32} 71.9 &  \cellcolor{blue!39} 77.7 &  \cellcolor{blue!38} 77.2 &  \cellcolor{blue!52} 87.8 &  \cellcolor{blue!60} 94.0 &  \cellcolor{blue!59} 93.6 \\
                      & sFLD &  \cellcolor{blue!10} 54.7 &  \cellcolor{blue!10} 54.5 &  \cellcolor{blue!10} 54.5 &  \cellcolor{blue!26} 67.9 &  \cellcolor{blue!21} 63.7 &  \cellcolor{blue!41} 79.1 \\
\midrule
                      & \textbf{avg.}        &  \cellcolor{white!20} 62.6 &  \cellcolor{white!40} 78.5 &  \cellcolor{white!40} 78.5 &  \cellcolor{white!38} 77.1 &  \cellcolor{white!35} 74.4 &  \cellcolor{white!43} \textbf{81.3} \\

\bottomrule
\end{tabular}
\vspace{-2mm}

\end{table}

We trained a prover on a deduction corpus (``source corpus'') and measured its performance on other corpora (``target corpus'') (\cref{tb:synthetic_transfer_arguments}).
The prover trained on \corpusSFLD performed the best on average, and as seen from the corpus-wise results, the prover transferred the most robustly to the other corpora while the provers trained on the other corpora did not exhibit this level of robustness.
Since the corpora used in \cref{tb:synthetic_transfer_arguments} differ in the set of deduction rules used in proofs, this result suggests that the prover trained \corpusSFLD generalized the most to other deduction rules.

The reason for this strongest generalizability should be the following.
\coolFont{(s)FLD} corpora teach LMs how to construct multistep deductions using the axioms.
Thanks to the completeness, the axioms can express multistep deductions constructed from any other deduction rules (including the ones used in the other corpora, as exemplified in \cref{appendix:argument:proofs}).
Thus, \textit{mastering the axioms leads to mastering various other deduction rules.}
On the other hand, the sets of deduction rules used in the other corpora do not have such a property and thus cannot be generalized to other deduction rules.

Since mastering various deduction rules is the most important in deductive reasoning, this generalizability to deduction rules obtained from \coolFont{FLD} corpora is vital.

\vspace{-3mm}
\subsection{Benchmarking by EntailmentBank}  \label{sec:NL_transfer}

\vspace{-4mm}
\begin{table}[h]
  \scriptsize
  \centering

\caption{
    The proof accuracy of provers on EntailmentBank.
    See \cref{appendix:sec:EB_other_metrics} for the results of other metrics.
   \label{tb:NL_transfer}
}
\vspace{-2mm}

\begin{tabular}{cllll}
\toprule

{} & {} & \multicolumn{3}{c}{EntailmentBank} \\
\cmidrule(l{\tabcolsep}r{\tabcolsep}){3-5}

{} & {} & Task1 & Task2 & Task3 \\
\midrule

  \multirow{3}{*}{\begin{tabular}[c]{@{}l@{}} Source \\ corpus\end{tabular}} &          T5 &  \cellcolor{blue!10} $36.8_{\pm 0.9}$ &  \cellcolor{blue!10} $31.2_{\pm 0.7}$ &   \cellcolor{blue!10} $6.2_{\pm 0.9}$ \\
         {} & RT.D5 &  \cellcolor{blue!60} $\mathrm{\mathbf{39.4}}_{\pm \mathrm{0.9}}$ &  \cellcolor{blue!38} $\mathrm{32.0}_{\pm \mathrm{0.8}}$ &   \cellcolor{blue!57} $\mathrm{8.2}_{\pm \mathrm{0.8}}$ \\
        {} & \textbf{FLD.D5} &  \cellcolor{blue!56} $\mathrm{39.2}_{\pm \mathrm{1.2}}$ &  \cellcolor{blue!60} $\mathrm{\mathbf{32.6}}_{\pm \mathrm{1.0}}$ &   \cellcolor{blue!60} $\mathrm{\mathbf{8.3}}_{\pm \mathrm{0.7}}$ \\
\bottomrule

\end{tabular}

\vspace{-3mm}
\end{table}

\Cref{tb:NL_transfer} shows the results on EntailmentBank (EB).
Since EB trees have high-depth (majority up to five), we used the high-depth versions of deduction corpora as source corpus.

First, as seen, the provers trained on both deduction corpora (\corpusRTDFiveWOS, \corpusFLDDFiveWOS) performed better than the baseline prover without such training (T5).
This suggests that the deductive reasoning ability acquired by synthetic deduction corpora generalizes to more complex real-world deductive reasoning.
We showcase some examples in \Cref{appendix:sec:EB_case_study}, where the error of the baseline prover is fixed by training on a deduction corpus (\corpusFLDDFiveWOS).
As seen, the prover captured the fundamentals of deduction rules better than the baseline as follows:
(i) it chose the correct premises necessary and sufficient to derive the next conclusion,
(ii) it included only such information that logically follows from the chosen premises into a conclusion, and
(iii) and it correctly used the rules of logical operators.

Looking at the results of deduction corpora closely, the prover trained on \corpusFLDDFive performed on par with the prover trained on \corpusRTDFiveWOS, even though it had mastered various deduction rules better, as shown in \cref{sec:effectiveness_of_FLD_arguments}.
We consider a possible reason as follows.
Firstly, real-world reasoning can require more coarse-grained deduction rules than those required by deduction corpora.
For expressing such coarse-grained deduction rules by the most fine-grained axioms, many steps are required, as in \cref{fig:proof_syllogism}.
However, the prover trained on \coolFont{FLD} still struggles with constructing many-step proofs using the axioms (detailed in \cref{sec:aspects_complexity_of_tree}).
In this sense, the prover could have failed to exploit the axioms' potential fully.
We will discuss future directions to tackle this challenge in \cref{sec:future_work}.

\vspace{-3mm}
\section{On What Aspects are Synthetic Deduction Corpora Beneficial?}    \label{sec:synthetic_transfer}
\vspace{-1mm}

A deduction corpus in \cref{tb:corpora} emphasizes a specific aspect different from those emphasized by the other corpora.
For each corpus (each aspect), we investigate whether the LM trained on that corpus outperforms the LM trained on the other corpus that does not emphasize the aspect.
If it does, we interpret it as meaning that the supervision from deduction corpus on that aspect is beneficial for LMs.

\vspace{-2mm}
\subsection{Ability to Solve Complex Proof Trees}    \label{sec:aspects_complexity_of_tree}
\vspace{-2mm}

\begin{table}[h]
\centering
\scriptsize
\tabcolsep 1.5mm

\caption{
    The depth-wise proof accuracies of the provers.
    \label{tb:synthetic_transfer_depth}
}
\vspace{-3mm}

    \begin{subfigure}{0.49\linewidth}
        \centering
        \subcaption{Target corpus is \corpusFLDRTOneWOS \label{tb:synthetic_transfer_depth_RT}}
        \begin{tabular}{llll}
        \toprule
        
        {} & {}  & \multicolumn{2}{c}{Source corpus} \\
        \cmidrule(l{\tabcolsep}r{\tabcolsep}){3-4}

        {} &                        T5 &                  FLD-impl.0 &                  FLD-impl.1 \\
        \midrule
        0 &  \cellcolor{blue!30} 41.7 &  \cellcolor{blue!51} 83.3 &  \cellcolor{blue!45} 70.8 \\
        1 &  \cellcolor{blue!48} 77.4 &  \cellcolor{blue!54} 88.7 &  \cellcolor{blue!56} 93.5 \\
        2 &  \cellcolor{blue!29} 38.0 &  \cellcolor{blue!38} 56.0 &  \cellcolor{blue!36} 53.0 \\
        3 &  \cellcolor{blue!26} 33.8 &  \cellcolor{blue!35} 50.0 &  \cellcolor{blue!33} 47.5 \\
        4 &  \cellcolor{blue!28} 36.7 &  \cellcolor{blue!35} 51.1 &  \cellcolor{blue!30} 40.0 \\
        5 &  \cellcolor{blue!20} 21.4 &  \cellcolor{blue!31} 42.9 &  \cellcolor{blue!31} 42.9 \\
        6 &  \cellcolor{blue!21} 22.7 &  \cellcolor{blue!29} 38.7 &  \cellcolor{blue!30} 41.3 \\
        7 &  \cellcolor{blue!22} 25.5 &  \cellcolor{blue!29} 38.7 &  \cellcolor{blue!32} 44.3 \\
        8 &  \cellcolor{blue!21} 23.0 &  \cellcolor{blue!28} 36.1 &  \cellcolor{blue!28} 37.7 \\
        \midrule
        \textbf{avg.} &  \cellcolor{white!27} 35.6 &  \cellcolor{white!36} 53.9 &  \cellcolor{white!36} 52.3 \\
        \bottomrule
        \end{tabular}
        
    \end{subfigure}
    \hfill
    \begin{subfigure}{0.49\linewidth}
        \centering
        \subcaption{Target corpus is \corpusFLDFourWOS  \label{tb:synthetic_transfer_depth_FLD}}
        \begin{tabular}{lll}
        \toprule
        
        {}  & \multicolumn{2}{c}{Source corpus} \\
        \cmidrule(l{\tabcolsep}r{\tabcolsep}){2-3}
        
                               T5 &                      FLD.3 &                      FLD.4 \\
        \midrule
         \cellcolor{blue!47} 75.0 &  \cellcolor{blue!60} 100.0 &  \cellcolor{blue!60} 100.0 \\
         \cellcolor{blue!42} 64.9 &   \cellcolor{blue!59} 98.6 &   \cellcolor{blue!57} 95.9 \\
          \cellcolor{blue!11} 2.3 &   \cellcolor{blue!41} 62.5 &   \cellcolor{blue!39} 59.1 \\
          \cellcolor{blue!10} 1.1 &   \cellcolor{blue!19} 18.9 &   \cellcolor{blue!21} 23.7 \\
          \cellcolor{blue!10} 0.0 &    \cellcolor{blue!10} 1.4 &    \cellcolor{blue!13} 7.5 \\
          \cellcolor{blue!10} 0.0 &    \cellcolor{blue!10} 0.9 &    \cellcolor{blue!11} 2.3 \\
          \cellcolor{blue!10} 0.0 &    \cellcolor{blue!10} 0.0 &    \cellcolor{blue!10} 0.0 \\
          \cellcolor{blue!10} 0.0 &    \cellcolor{blue!10} 0.0 &    \cellcolor{blue!10} 0.0 \\
          \cellcolor{blue!10} 0.0 &    \cellcolor{blue!10} 0.0 &    \cellcolor{blue!10} 1.2 \\
          \midrule
          \cellcolor{white!17} 15.9 &   \cellcolor{white!25} 31.4 &   \cellcolor{white!26} 32.2 \\
        \bottomrule
        \end{tabular}

    \end{subfigure}

\vspace{-2mm}

\end{table}

\Cref{tb:synthetic_transfer_depth} shows the depth-wise performances of provers.
The corpora in \Cref{tb:synthetic_transfer_depth_RT} use the implication deduction rules.
The prover trained on the corpus of shallower ($\sim 3$) trees (\corpusFLDRTZeroWOS) generalizes to deeper ($4 \sim 8$) trees to some extent, and performs similarly to the prover trained on the corpus of deeper trees (\corpusFLDRTOneWOS).
This generalization to deeper trees coincides with previous findings \cite{tafjord-etal-2021-proofwriter,sanyal2022fairr}.
However, as \cref{tb:synthetic_transfer_depth_FLD} shows, when the corpora use the axioms, neither the provers trained on the shallower tree corpus (\corpusFLDThreeWOS) nor deeper tree corpus (\corpusFLDFourWOS) failed in solving deeper trees.

We can interpret this seemingly contradictory result as follows.
As discussed in \cref{sec:how_well}, the number of possible proof tree patterns can be estimated (very roughly) as $\mathcal{O}(\mathcal{A}^d)$.
When a prover tries to solve a deduction example, it has to choose and generate exactly the one gold proof tree from these possible negative proof trees.
This should be very difficult for large $d$ with large $\mathcal{A}$.
Now, while the corpora in \cref{tb:synthetic_transfer_depth_RT} use a few deduction rules ($\mathcal{A}=2$) of implication type, corpora in \cref{tb:synthetic_transfer_depth_FLD} use various deduction rules ($\mathcal{A} \sim 10$) of the axioms.
This made it very difficult to solve large-depth deduction examples of these corpora, which lead the provers to fail in solving large-depth proof trees in \cref{tb:synthetic_transfer_depth_FLD}.

Overall, for solving complex trees, the supervision from deduction corpora can be necessary but not sufficient alone.

\vspace{-2mm}

\subsection{Understanding of Diverse Linguistic Expressions}   \label{sec:linguistic_diversity}

\begin{table}[h]
\centering
\scriptsize

\caption{
    \captionFewShot{the diversity of linguistic expressions}
    \label{tb:synthetic_transfer_NL}
}
\vspace{-3mm}

\begin{tabular}{lllllll}
\toprule

{} & {}  & {} & \multicolumn{4}{c}{Source corpus} \\
\cmidrule(l{\tabcolsep}r{\tabcolsep}){4-7}

{} & {} & {} & \multicolumn{2}{c}{RuleTaker} & \multicolumn{2}{c}{FLD} \\
\cmidrule(l{\tabcolsep}r{\tabcolsep}){4-5} \cmidrule(l{\tabcolsep}){6-7}

{} & {} &                        T5 &                        RT &                     RT.PR &                     FLD.2 &                     FLD.3 \\
\midrule
\multirow{4}{*}{\begin{tabular}[c]{@{}l@{}} Target \\ corpus\end{tabular}} & RT    &  \cellcolor{blue!42} 70.1 &  \cellcolor{blue!58} 92.4 &  \cellcolor{blue!58} 91.3 &  \cellcolor{blue!48} 78.3 &  \cellcolor{blue!47} 76.6 \\
                                                                                    & RT.BE &  \cellcolor{blue!38} 64.3 &  \cellcolor{blue!58} 91.3 &  \cellcolor{blue!60} 93.9 &  \cellcolor{blue!43} 71.3 &  \cellcolor{blue!45} 73.4 \\
                                                                                    & FLD.2 &  \cellcolor{blue!14} 31.0 &  \cellcolor{blue!16} 34.2 &  \cellcolor{blue!17} 34.7 &  \cellcolor{blue!40} 66.8 &  \cellcolor{blue!39} 66.2 \\
                                                                                    & FLD.3 &  \cellcolor{blue!10} 24.8 &  \cellcolor{blue!12} 28.7 &  \cellcolor{blue!11} 27.5 &  \cellcolor{blue!39} 65.3 &  \cellcolor{blue!40} 66.4 \\
                                                                                    \midrule
                                                                                    {} & \textbf{avg.}  &  \cellcolor{white!26} 47.6 &  \cellcolor{white!36} 61.6 &  \cellcolor{white!36} 61.8 &  \cellcolor{white!42} 70.4 &  \cellcolor{white!43} 70.7 \\

\bottomrule
\end{tabular}

\vspace{-3mm}
\end{table}

\Cref{tb:synthetic_transfer_NL} shows that a prover trained on a corpus with less linguistic diversity (i.e., \corpusRT and \corpusFLDTwoWOS) performed as well as the prover trained on the linguistically diverse counterpart of that corpus (i.e., \corpusRTNL and \corpusFLDThreeWOS, respectively).
This suggests that LMs are self-sufficient on the linguistic aspect, and thus additional supervision from deduction corpora is not that important.

Indeed, this result coincides with the previous findings \cite{clark2020transformers,tafjord-etal-2021-proofwriter} and can be intuitively understood: since the pre-training corpora of LMs are huge and linguistically diverse, they should have given LMs many chances to learn linguistic of logical statements such as that ``If A, then B'' paraphrases to ``A leads to B''.

\subsection{Understanding of Complex Formulas}   \label{sec:aspects_complexity_of_formula}

\begin{table}[h]
\centering
\scriptsize

\caption{
    \captionFewShot{the complexity of formulas}
    \label{tb:synthetic_transfer_logical_formula_complexity}
}
\vspace{-2mm}

\begin{tabular}{lllll}
\toprule

{} & {}  & {} & \multicolumn{2}{c}{Source corpus} \\
\cmidrule(l{\tabcolsep}r{\tabcolsep}){4-5}

{} & {} &                        T5 &                     FLD.1 &                     FLD.2 \\
\midrule
\multirow{2}{*}{\begin{tabular}[c]{@{}l@{}} Target \\ corpus\end{tabular}} & FLD.1 &  \cellcolor{blue!22} 43.1 &  \cellcolor{blue!60} 77.9 &  \cellcolor{blue!53} 71.6 \\
                                                                            & FLD.2 &  \cellcolor{blue!10} 31.0 &  \cellcolor{blue!25} 46.0 &  \cellcolor{blue!48} 66.8 \\
\bottomrule
\end{tabular}

\end{table}
\vspace{-2mm}

\Cref{tb:synthetic_transfer_logical_formula_complexity} shows that while the prover trained on the corpus with simple formulas (\corpusFLDOneWOS) performed poorly on the corpus with complex formulas (\corpusFLDTwoWOS), the prover trained \corpusFLDTwo performed well on both corpora.
Thus, deduction corpora are beneficial for mastering complex formulas.

We can interpret this result as follows.
The complex formulas included in \corpusFLDTwo are formed by modifying atomic formulas with logical operators $\lnot, \land, \lor$.
The semantics of these logical operators, such that ``a sentence with negation $\lnot$ have the opposite meaning of that sentence without negation'', and that ``$A \lor B$ does not necessarily imply $A$'', are seldom written explicitly by humans.
Thus, the pre-training corpora gave LMs too few chances for learning these semantics.
This result is enhanced by the previous findings that LMs fail to understand the semantics of negation \cite{naik-etal-2018-stress,hossain-etal-2020-analysis,kassner-schutze-2020-negated}.

\vspace{-2mm}
\subsection{Robustness to Distractive Facts}    \label{sec:aspects_distractor}
\vspace{-2mm}

\begin{table}[H]
\centering
\scriptsize

\caption{
    \captionFewShot{the number of distractors}
    \label{tb:synthetic_transfer_distractors}
}
\vspace{-3mm}

\begin{tabular}{lllll}
\toprule

{} & {}  & {} & \multicolumn{2}{c}{Source corpus} \\
\cmidrule(l{\tabcolsep}r{\tabcolsep}){4-5}

{} & {} &                        T5 &                     FLD.0 &                     FLD.2 \\
\midrule
\multirow{2}{*}{\begin{tabular}[c]{@{}l@{}} Target \\ corpus\end{tabular}} & FLD.0 &  \cellcolor{blue!18} 38.9 &  \cellcolor{blue!60} 76.4 &  \cellcolor{blue!58} 75.2 \\
                                                                           & FLD.2 &  \cellcolor{blue!10} 31.0 &  \cellcolor{blue!38} 56.7 &  \cellcolor{blue!49} 66.8 \\
\bottomrule
\end{tabular}

\vspace{-5mm}
\end{table}

\Cref{tb:synthetic_transfer_distractors} shows that, while the prover trained on the corpus without distractors (\corpusFLDZeroWOS) performed poorly on the corpus with distractors (\corpusFLDTwoWOS), the prover trained on \corpusFLDTwo performed well on both corpora.
Thus, synthetic distractors are beneficial for acquiring the robustness to distractive facts.
This result is intuitive: since the human-written text should not include the facts irrelevant to the content, the pre-training corpora should not have given LMs a chance to acquire robustness to irrelevant facts.

\vspace{-2mm}
\section{Discussions and Future Directions}   \label{sec:future_work}
\vspace{-1mm}

So far, we have investigated each aspect of deductive reasoning.
We summarize the results and discuss future directions.

\noindent \textbf{Mastery on Various Deduction Rules: }
Mastering various deduction rules is the most important in deductive reasoning.
We showed that \coolFont{FLD} corpora teach LMs various deduction rules the most effectively (\cref{sec:effectiveness_of_FLD_arguments}).
This should be because that \coolFont{FLD} adopts the axioms of first-order predicate logic system, which can derive any valid deduction rules in this system.
The next step will be to examine the axioms of other logic systems, such as linear and modal logic systems, which are also important in real-world reasoning.

\vspace{-1mm}
\noindent \textbf{Ability to Solve Complex Proof Trees: }
We have shown that solving a many-step proof tree is still challenging for LMs even after training on deduction corpora (\cref{sec:aspects_complexity_of_tree}).
The possible reason is that they have to choose and generate a gold proof from a large number of possible trees.
To solve this problem, inventing smarter and strategic search methods on possible generation space, such as \citet{li2016deep,negrinho2018learning,picco-etal-2021-neural-unification,welleck2022naturalprover}, could be a promising direction.

\vspace{-1mm}
\noindent \textbf{Understanding of Complex Formulas: }
We have shown that deduction corpora are effective for LMs to understand the semantics of logical operators such as $\neg, \land, \lor$ (\cref{sec:aspects_complexity_of_formula,sec:NL_transfer}).
It could be even more effective to incorporate the recent learning methodological approaches for making LMs understand negation \cite{prollochs-etal-2019-learning,hosseini-etal-2021-understanding} into the learning on deduction corpora.

\vspace{-1mm}
\noindent \textbf{Robustness to Distractive Facts: }
We have shown that the synthetic distractors can make LMs robust to distractive facts (\cref{sec:aspects_distractor}).
In a real scenario of logical reasoning, the facts have to be collected by possibly incomplete retrieval systems.
The distractors that imitate ones appearing in such a scenario could be more effective.
We can generate such distractors as follows:
(i) We build a database of synthetic facts.
(ii) For a given deduction example, we collect facts from the database by actual retrieval systems.

\vspace{-1mm}
\noindent \textbf{Generalization to Real-World Reasoning Tasks: }
We have shown that the training on deduction corpora is even useful for deductive reasoning in a more real-world setting (\cref{sec:NL_transfer}).
However, the LMs trained on \coolFont{FLD} could not fully utilize the potential of the axioms, as they failed in constructing many-step proofs to express coarse-grained deduction rules, which could be required in real-world reasoning (\cref{sec:NL_transfer,sec:aspects_complexity_of_tree}).
We discussed future directions to solve such many-step proofs above.

Further, LMs may need additional training to utilize deduction rules well in a realistic context.
For example, the LMs could have to combine deduction rules with common sense knowledge, use multiple deduction rules at once to jump to the next conclusion, and judge the validity of a proof step considering the overall context.
Recently, \citet{wei2022chain,kojima-2022} showed that large LMs can utilize deduction rules in a realistic context, given appropriate prompts.
It could be promising to integrate this approach and deduction corpora training.

Pursuing further real-world scenarios, we have to tackle tasks of other settings.
One is deductive reasoning that requires us to collect relevant facts by ourselves.
For this, we could exploit factual knowledge implicitly embedded in LMs \cite{petroni-etal-2019-language,davison-etal-2019-commonsense,talmor2020leap}, or use retrieval systems.
For the latter, we could train LM-based retrievers \cite{karpukhin2020dense,guu2020retrieval} using synthetic deduction examples and fact database.
Abductive reasoning \cite{bhagavatula2019abductive} is another kind of real-world logical reasoning with which we derive hidden premises from a conclusion and other visible premises.
Synthetic corpora for abduction based on formal logic can be generated similarly to as done in this study.

\vspace{-3mm}
\section{Conclusion}   \label{sec:conclusion}
\vspace{-2mm}

To teach language models deductive reasoning, we proposed a synthetic corpus based on formal logic theory and verified its effectiveness empirically.
Further, we analyzed each aspect of deductive reasoning and provided future directions on each.
We will advance on the basis of these directions.

\section*{Acknowledgement}
We thank the three anonymous reviewers and the meta-reviewer, who gave us insightful comments and suggestions.
Computational resources of AI Bridging Cloud Infrastructure (ABCI) provided by the National Institute of Advanced Industrial Science and Technology (AIST) were used.
We thank Dr. Masaaki Shimizu at Hitachi for the convenience of additional computational resources.
We thank Dr. Naoaki Okazaki, professor at Tokyo Institute of Technology, for the keen comments.

\bibliography{icml2023}
\bibliographystyle{icml2023}

\appendix

\numberwithin{equation}{section}
\renewcommand{\thefigure}{\Alph{section}.\arabic{figure}}
\renewcommand{\thetable}{\Alph{section}.\arabic{table}}

\newpage
\clearpage

\section{Related Work} \label{appendix:sec:related_works}

\subsection{Synthetic Corpus for Acquiring Deductive Reasoning Ability}

A synthetic deduction corpus can be one promising approach for language models (LMs) to acquire logical deductive reasoning ability.
The automatic (programmatic) generation ensures the validity of the resulting deductive proof examples.
Further, since we can bypass high-cost human annotations we can generate many examples, which should be required by LMs to learn deductive reasoning inductively.

RuleTaker \cite{clark2020transformers} proposed a deduction corpus composed of synthetically generated multistep deductive proofs written in natural languages.
Each deductive proof (dis-)proves a hypothesis by applying deduction rules multiple times to a given set of facts.
As the deduction rules, rules of the implication kind were used.
They showed that Transformer \cite{vaswani2017attention} LMs can solve these problems in the sense that they can predict the final answer (i.e., ``proved'', ``disproved'', or ``unknown'') of each deductive proof given the fact set.
Later studies \cite{saha-etal-2020-prover,dalvi-etal-2021-explaining,tafjord-etal-2021-proofwriter,sanyal2022fairr} showed that generative LMs can generate even the intermediate proofs as well as the final answer.

Artificial Argument Corpus \cite{betz-etal-2021-critical} is another corpus composed of synthetically generated single-step deductive proofs.
As the deduction rules, hand-selected rules useful for critical thinking were used.
They showed that the LMs trained on this corpus not only solve the task of this corpus itself but generalize to other NLI tasks from GLUE benchmark \cite{wang-etal-2018-glue}.
However, at the same time, they showed that such LMs do not generalize well to more challenging logical reasoning tasks such as ARC \cite{habernal-etal-2018-argument} and LogiQA \cite{ijcai2020p501}.

\citet{gontier2020measuring} investigated the deductive reasoning ability of LMs on a corpus, which is composed of a specific type of multistep inference: kinship relationship on synthetic kinship graphs.
They found that LMs can solve this task when the number of proof steps is not large, but it is difficult for them to generalize to longer-than-training proofs.
\citet{bostrom-etal-2021-flexible} studied how we can create realistic natural language expressions that represent deduction rules.
To this end, they scraped sentences from Wikipedia by a template-based method and paraphrased them.
They showed that training on this corpus is helpful for solving real-world deductive reasoning such as EntailmentBank \cite{dalvi-etal-2021-explaining}.

While all these corpora focused on specific sets of deduction rules, we focus on the theoretically well-grounded set of deduction rules that can derive any other deduction rules.
Further, we analyze each aspect of deductive reasoning using corpora of various patterns to advance the research direction of deductive reasoning.

\subsection{Benchmarks for Deductive Reasoning}

Many benchmarks of single-step logical reasoning using specific reasoning rules have been proposed: bAbI \cite{weston2015towards}, QuaRTz \cite{tafjord2019quartz}, ROPES \cite{lin-etal-2019-reasoning} and \citet{richardson2020probing}.
For multistep deductive reasoning, FOLIO \cite{han-2022} is a human-authored benchmark of the SAT (i.e., satisfiability) problem.
Given a set of facts and hypotheses, which are created by a human referencing a specific page of Wikipedia, we are required to assign a truth value to each hypothesis.
This requires (implicitly) conducting multistep deductive reasoning using high-granularity deduction rules.

RuleTaker \cite{clark2020transformers,tafjord-etal-2021-proofwriter} can work as a benchmark as well as a synthetic training corpus.
RuleTaker focuses on multistep deductive reasoning constructed from implication rules.
Since RuleTaker requires generating all the intermediate steps as well as the final prediction on the hypothesis, it is suitable for measuring deductive reasoning ability explicitly and transparently.
Further, it includes many irrelevant facts so that the model has to choose only relevant facts under these noises.
This makes the task challenging.
LogicNLI \cite{tian2021diagnosing} can be considered as an extension of RuleTaker, where additional logical operators such as ``$\equiv$'' are used
\footnote{
We do not compare our method with this study directly because the code and corpora are not publicly available and the paper does not clarify the exact rules used.
}
.
Additionally, the examples of LogicNLI are checked manually by humans to ensure their quality.

EntailmentBank (EB) \cite{dalvi-etal-2021-explaining} is the same type of task as RuleTaker, but is even more challenging.
The proof trees in EB dataset are human-authored rather than synthetically generated.
Further, each proof step can be a rough entailment instead of a rigid logical step.
Thus, EB measures logical reasoning ability in a more real-world setting.

\subsection{Proof Generation Models}

Earlier work \cite{saha-etal-2020-prover,gontier2020measuring,dalvi-etal-2021-explaining,sun-etal-2021-probabilistic} generated proof sequences at once by LMs.
Later work \cite{tafjord-etal-2021-proofwriter,bostrom-2022,sanyal2022fairr,yang2022nlproofs} generated proofs step-wisely one by one.
The stepwise methods are more faithful and robustly generalize to longer proofs.
Recently, \citet{wei2022chain,kojima-2022} showed that large language models (LLMs) perform well on multi-hop inference tasks provided appropriate prompts.
However, \citet{yang2022nlproofs} showed that LLMs given few-shot examples perform poorer than fine-tuned smaller LMs.

The synthetic corpora approach examined in this paper could potentially help all these models to acquire better deductive reasoning ability.

\section{Limitations}  \label{appendix:sec:limitations}
The study has the following limitations:
\begin{itemize}
    \item We examined only a kind of logical reasoning: deductive reasoning with a given set of facts. As stated in \cref{sec:future_work}, we have other types of logical reasoning to be solved in the future.
    \item We examined only the first-order predicate logic system, while there are other logic systems useful for real-world reasoning to be tackled in the future, as stated in \cref{sec:future_work}.
\end{itemize}

\begin{figure*}[t!]
    \centering
        \begin{subfigure}[t]{1.0\linewidth}
        \centering
        \includegraphics[width=1.0\linewidth]{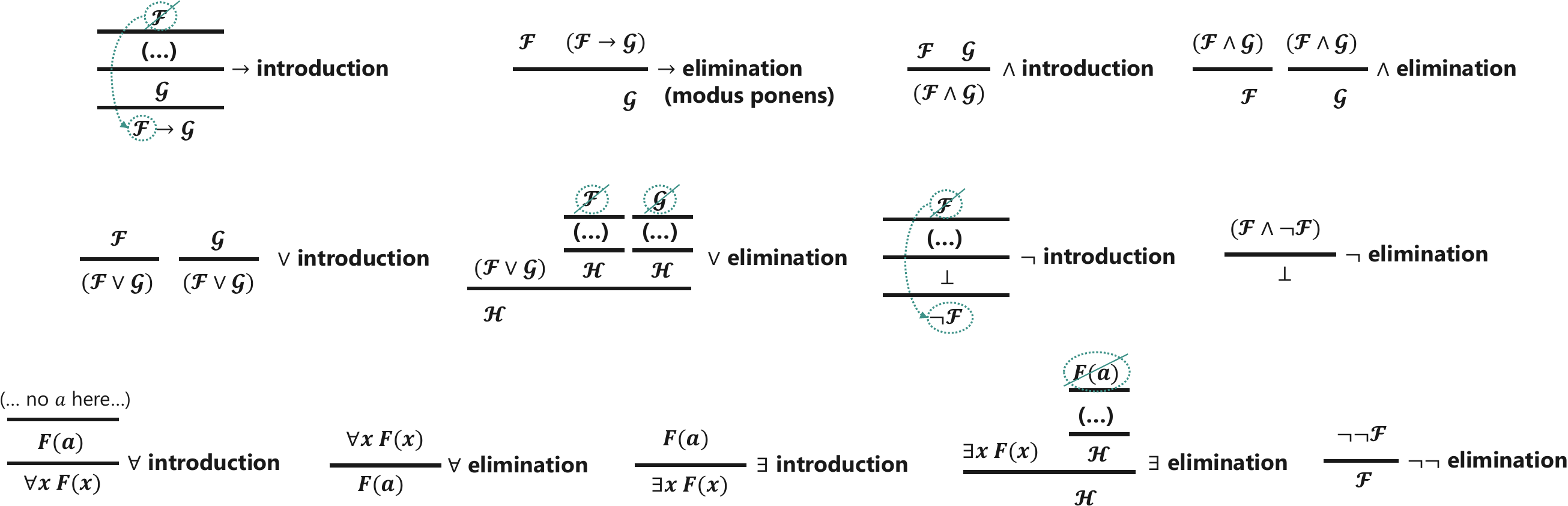}
        \subcaption{
        The axioms of first-order predicate logic used in FLD.\label{appendix:argument:axioms}
        }
    \end{subfigure}
    
    \vspace{5mm}
    \begin{subfigure}[t]{0.5\linewidth}
            \centering
            \includegraphics[width=0.75\linewidth]{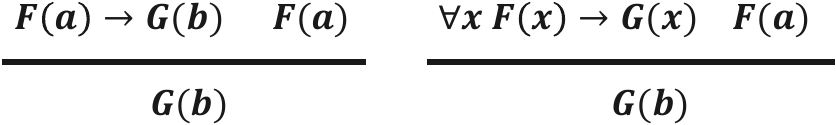}
            \subcaption{
            The deduction rules of the implication type used in RuleTaker \cite{clark2020transformers}.\label{appendix:argument:implication}
            }
    \end{subfigure}
    \caption{
    We show the deduction rules used in relevant corpora.
    For the ``critical thinking'' deduction rules used in AACorpus \cite{betz-etal-2021-critical}, please refer to Figure 1 in \citet{betz-etal-2021-critical}.
    \label{appendix:arguments:all}
    }
\end{figure*}

\begin{figure}[t!]
    \centering
    \begin{subfigure}{0.48\linewidth}
        \centering
        \includegraphics[width=1.0\linewidth]{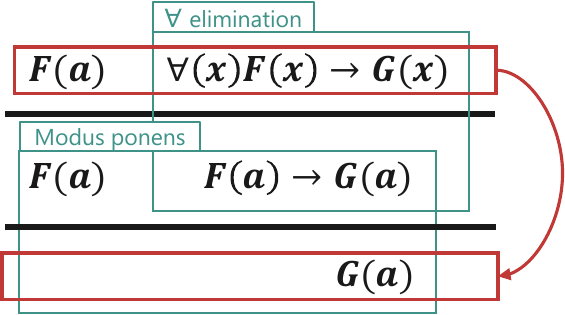}
        \subcaption{
            Derivation of a deduction rule of implication type used in RuleTaker \cite{clark2020transformers}.
            \label{appendix:argument:proof_G_modus_ponens}
        }
    \end{subfigure}
    \hfill
    \begin{subfigure}{0.48\linewidth}
        \centering
        \includegraphics[width=1.0\linewidth]{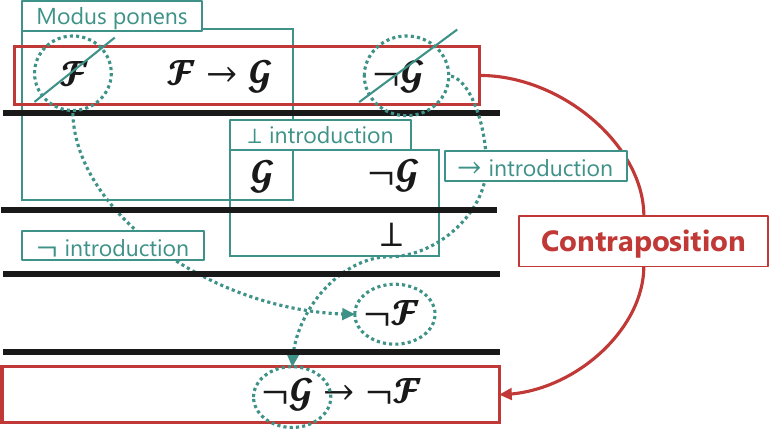}
        \subcaption{
            Derivation of contraposition deduction rule used in AACorpus \cite{betz-etal-2021-critical}.
            \label{appendix:argument:proof_contraposition}
        }
    \end{subfigure}
    \caption{As examples of multistep deduction constructed from the axioms, we show the derivations of the deduction rules used in the previous studies. \label{appendix:argument:proofs}}
\end{figure}

\begin{table}[t!]
    \scriptsize
    \caption{
        Truth values of the premises ($\mathcal{P}_i$) and the conclusion ($\mathcal{C}$) of the two deduction rules.
        \blue{Blue} shows truth value assignments where both the premises and the conclusion are true (=1).
        \red{Red} shows truth value assignments where the conclusion is false even if all the premises are true.
        \label{tb:truth_tables}
    }
    
    \begin{subfigure}{0.55\linewidth}
        \centering
        \subcaption{
            The valid deduction rule \cref{eq:argument:syllogism} {\scriptsize(syllogism)}.\\
            $\mathcal{P}_1=(\mathcal{F} \rightarrow \mathcal{G}) \land (\mathcal{G} \rightarrow \mathcal{H})$,\\
            $\mathcal{C}=\mathcal{F} \rightarrow \mathcal{H}$.
            \label{tb:truth_table_syllogism}
        }
                
        \begin{tabular}{@{}lll|ll@{}}
        \toprule
        $\mathcal{F}$ & $\mathcal{G}$ & $\mathcal{H}$ & $\mathcal{P}_1$ & $\mathcal{C}$ \\ \midrule
        1             & 1             & 1             & \blue{1}        & \blue{1}      \\
        1             & 1             & 0             & 0               & 0             \\
        1             & 0             & 1             & 0               & 1             \\
        1             & 0             & 0             & 0               & 0             \\
        0             & 1             & 1             & \blue{1}        & \blue{1}      \\
        0             & 1             & 0             & 0               & 1             \\
        0             & 0             & 1             & \blue{1}        & \blue{1}      \\
        0             & 0             & 0             & \blue{1}        & \blue{1}      \\ \bottomrule
        \end{tabular}
    \end{subfigure}
    \hspace{1mm}
    \begin{subfigure}{0.42\linewidth}
        \centering
        \subcaption{
            The invalid deduction rule \cref{eq:argument:incorrect}.\\
            $\mathcal{P}_1=\mathcal{F},\mathcal{P}_2=\mathcal{F}\lor \mathcal{G}$,\\
            $\mathcal{C}=\mathcal{G}$.
            \label{tb:truth_table_wrong}
        }
        
        \begin{tabular}{@{}ll|lll@{}}
        \toprule
        $\mathcal{F}$ & $\mathcal{G}$ & $\mathcal{P}_1$ & $\mathcal{P}_2$ & $\mathcal{C}$ \\ \midrule
        1             & 1             & \blue{1}        & \blue{1}        & \blue{1}      \\
        1             & 0             & \red{1}         & \red{1}         & \red{0}       \\
        0             & 1             & 0               & 1               & 1             \\
        0             & 0             & 0               & 0               & 0             \\ \bottomrule
        \end{tabular}
    \end{subfigure}
   
\end{table}

\section{Ethics and Social Impacts}  \label{appendix:sec:social_impacts}
The ultimate goal of the direction of this study is to make an artificial intelligence (AI) that reasons logically step by step.
If AI can make a decision by showing logical steps one by one, it will be highly explainable and transparent to users.
Furthermore, a user will be able to trace the error of AI.
Thus, we think this study makes steps towards AI that will positively impact society.

\begin{figure*}[t!]
    \centering
        \begin{subfigure}[t]{1.0\linewidth}
        \centering
        \includegraphics[width=1.0\linewidth]{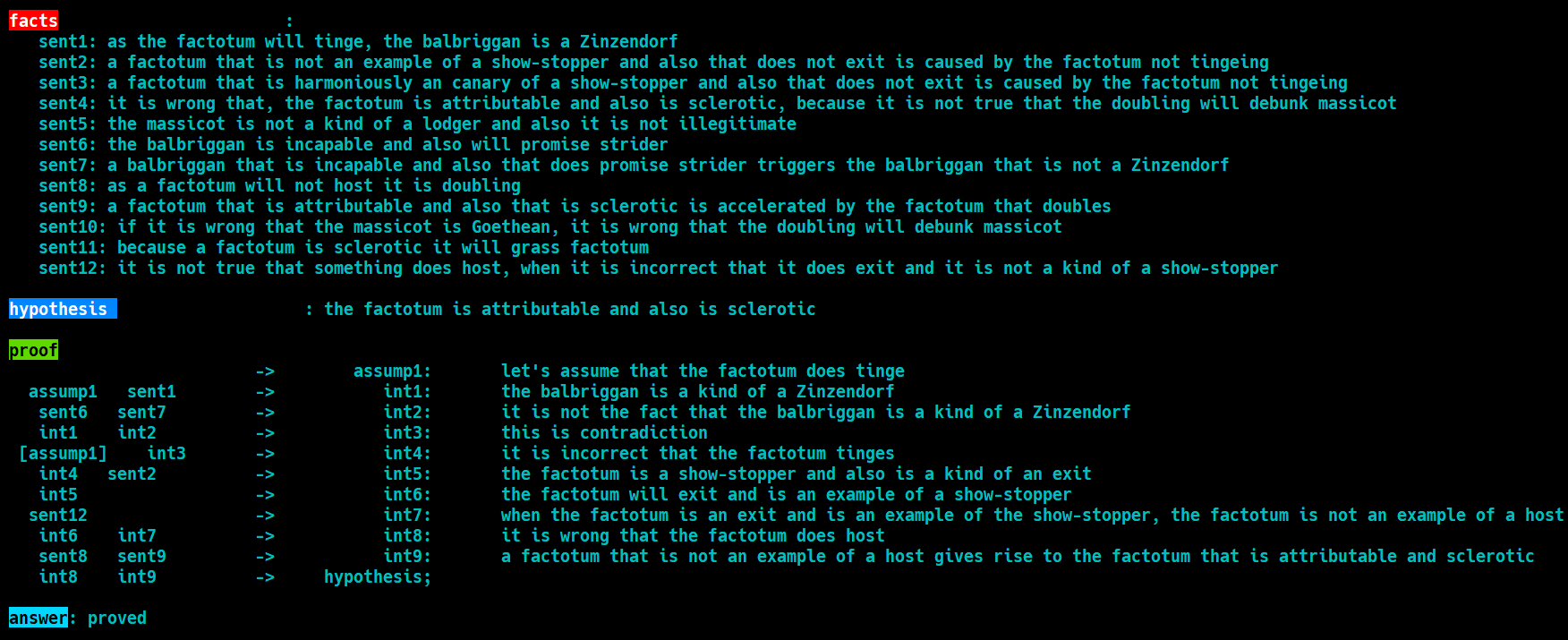}
        \subcaption{
        An example where the proof by contradiction is used. \label{aappendix:fig:FLD_output_sample_proof_by_contradiction}
        }
    \end{subfigure}
    
    \vspace{5mm}
    
    \centering
        \begin{subfigure}[b]{0.80\linewidth}
        \centering
        \includegraphics[width=1.0\linewidth]{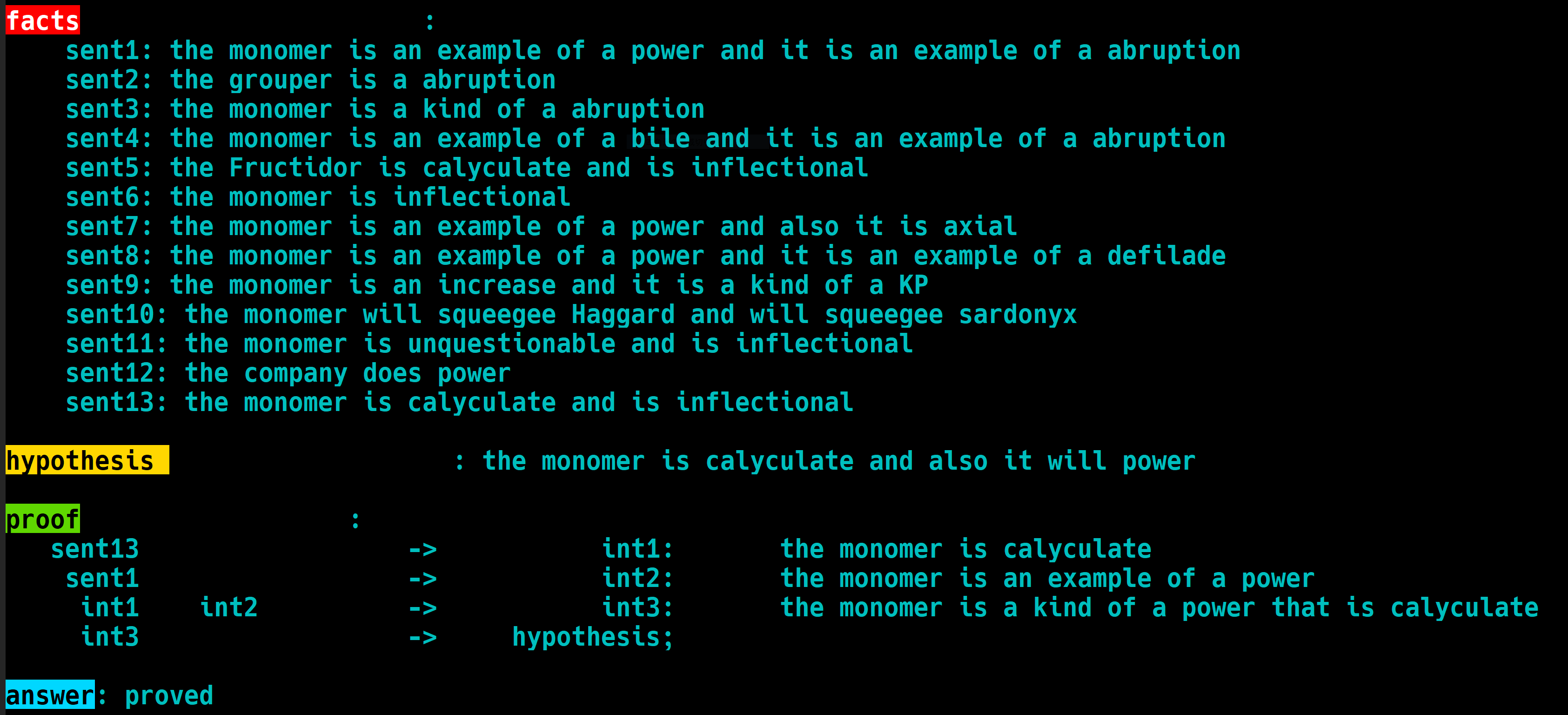}
        \subcaption{
        An example that can test the semantics of logical conjunction $\land$. \label{aappendix:fig:FLD_output_sample_and}
        }
    \end{subfigure}
    \caption{
    Examples of FLD examples.
    \label{appendix:fig:FLD_output_sample}
    }
\end{figure*}

\section{Formal Logic} \label{appendix:sec:formal_logic}

\subsection{Definition of Validity of a Deduction Rule}

A deduction rule is valid when for all truth value assignments, the conclusion is always true (=1) if all the premises are true (=1).
This is exemplified in \cref{tb:truth_table_syllogism}

A deduction rule is invalid when for some truth value assignments, the conclusion is false (=0) even if all the premises are true (=1).
This is exemplified in \cref{tb:truth_table_wrong}.

\subsection{Deductoin Rules used in Relevant Corpora}

\Cref{appendix:argument:axioms} shows the axioms of first-order predicate logic.
\Cref{appendix:argument:implication} shows the deduction rules of implication type used in RuleTaker \cite{clark2020transformers}.
For the deductoin rules used in AACorpus, see Figure 1 in \citet{betz-etal-2021-critical}.

\subsection{Examples of Multistep Deduction constructed from the Axioms}
\Cref{appendix:argument:proof_G_modus_ponens} shows the derivation of a deduction rule of implication type used in RuleTaker.
\Cref{appendix:argument:proof_contraposition} shows the derivation of the contraposition deduction rule used in AACorpus.

\section{Corpus Generation based on Formal Logic Deduction}   \label{appendix:sec:FLD}

\begin{figure}[t!]
    \centering
        \includegraphics[width=0.75\linewidth]{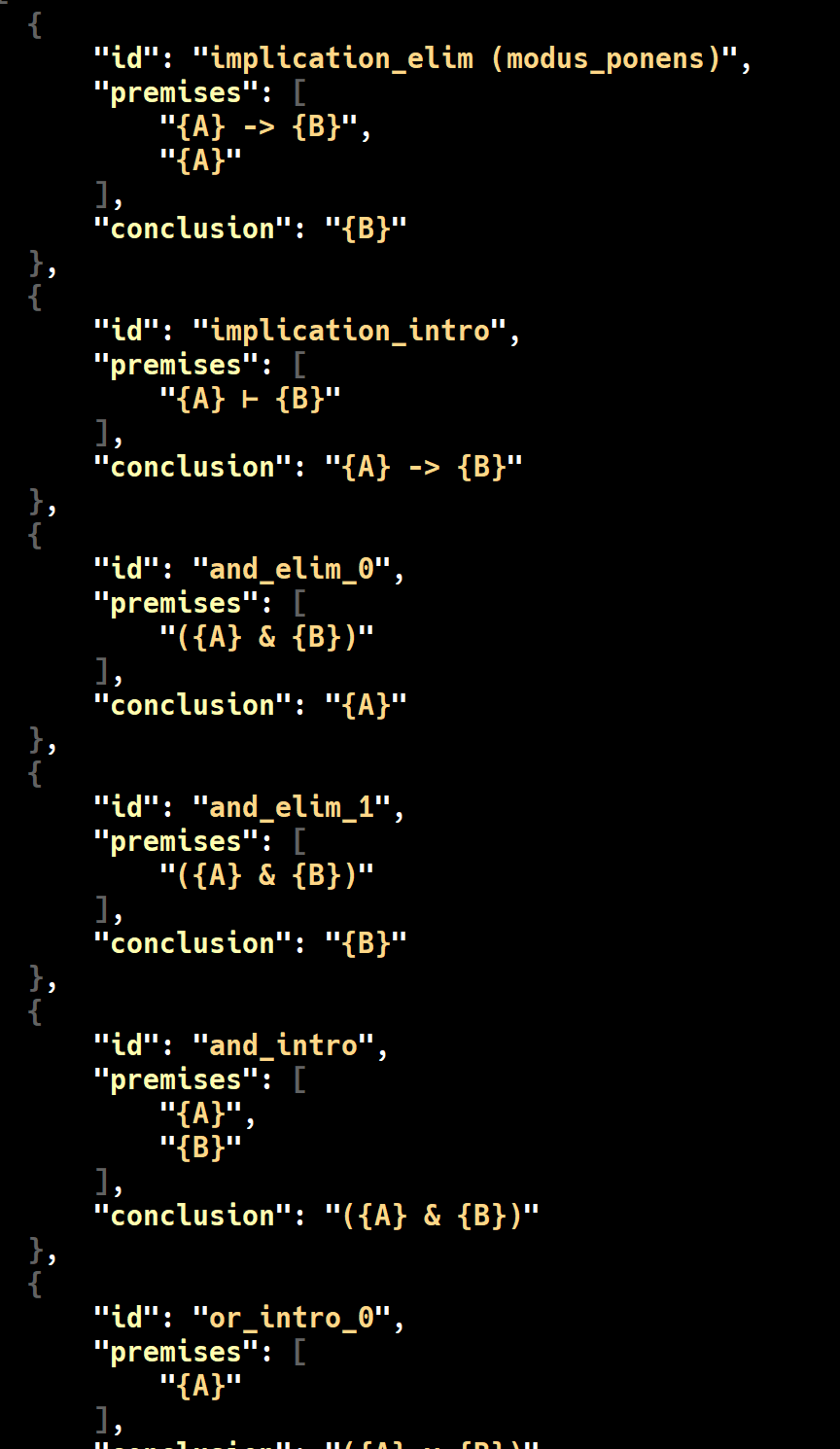}
    \caption{
    An example of the template rule file for deduction rules used by the proof tree generator.
     \label{appendix:fig:argument_template}
    }
\end{figure}

\begin{figure*}[t!]
    \centering
        \begin{subfigure}[t]{0.7\linewidth}
        \centering
        \includegraphics[width=1.0\linewidth]{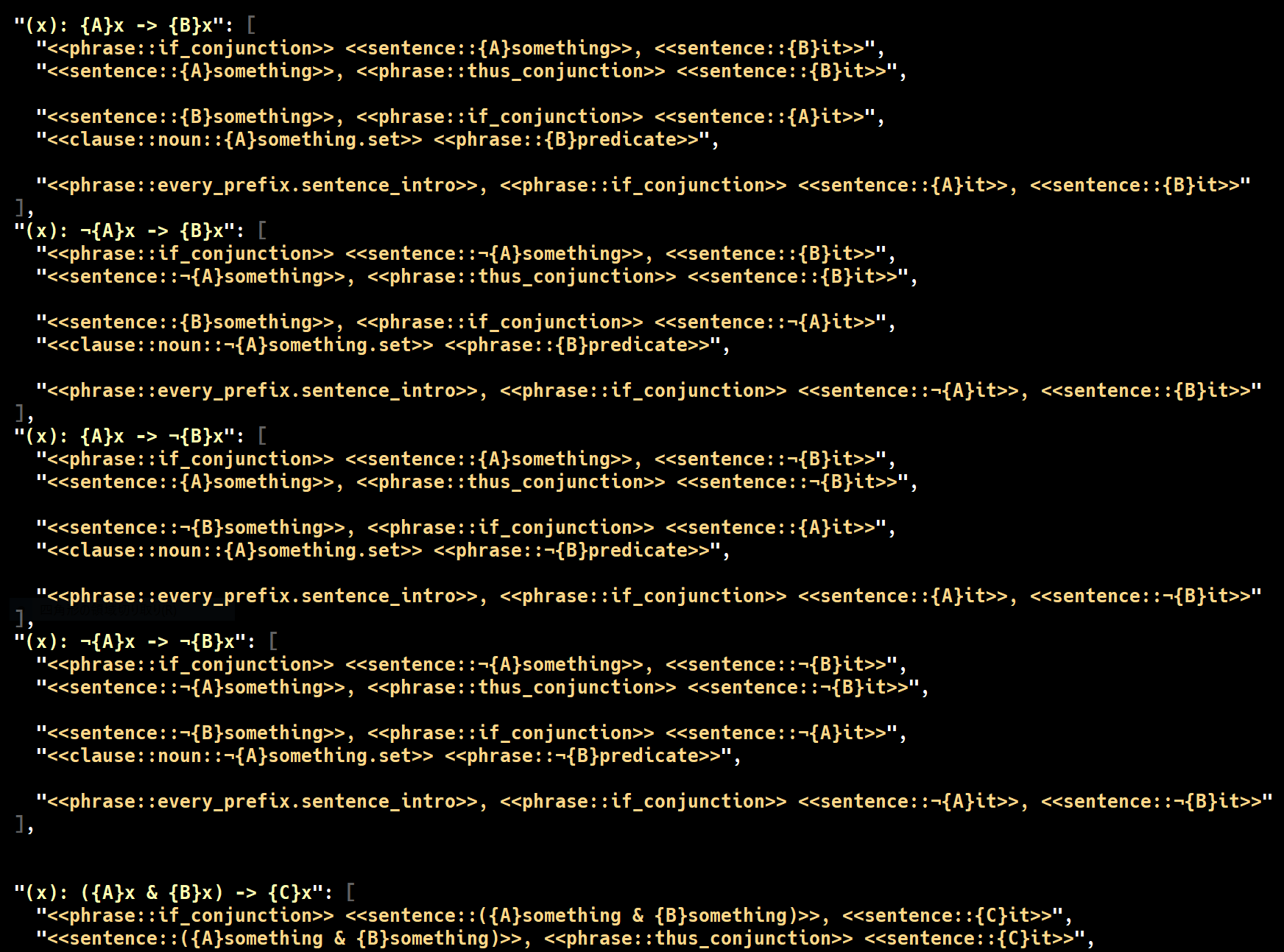}
        \subcaption{
        For each formula, we have several templates.
        A template can be nested (redirected to other templates), as shown. \label{appendix:fig:translation_template_0}
        }
    \end{subfigure}
    \hfill
    \centering
        \begin{subfigure}[b]{0.29\linewidth}
        \centering
        \includegraphics[width=1.0\linewidth]{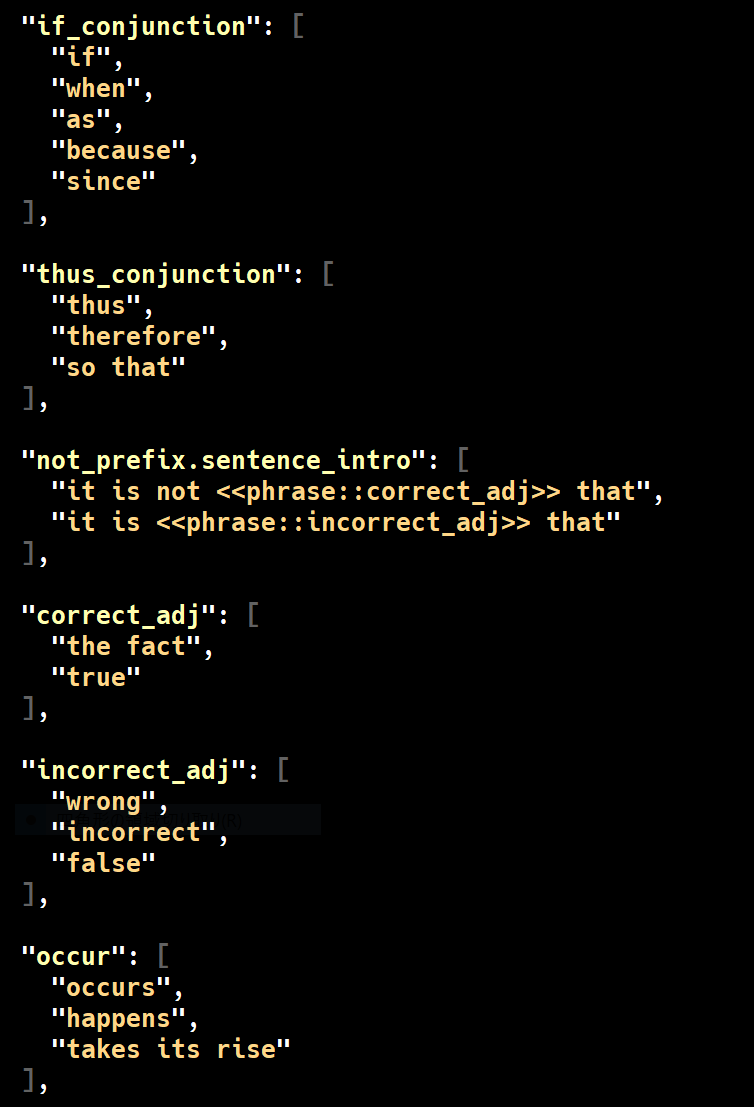}
        \subcaption{
        The final natural language templates of logical phrases. \label{appendix:fig:translation_template_1}
        }
    \end{subfigure}
    \caption{
    An example the template file of the natural language assigner
     \label{appendix:fig:translation_template}
    }
\end{figure*}

We show the examples of FLD example in \cref{appendix:fig:FLD_output_sample}

Below, we show additional details of each module of \coolFont{FLD}.
Please refer to \cref{fig:framework_overview} on intuitive understanding.

\subsection{Proof Tree Generation}   \label{appendix:sec:FLD_proof_tree_generation}

As stated in \cref{sec:FLD_proof_tree_generation}, the generator module generates a proof tree by forward- and backward random deduction, using the deduction rules specified by a user.
A user can specify deduction rules via a template file as exemplified in \cref{appendix:fig:argument_template}.

The forward random deduction is done as follows.
The generator first chooses a deduction rule randomly and forms the initial tree where the root node is the conclusion of the chosen deduction rules and the child nodes are the premises of the chosen deduction rule.
The generator next randomly chooses another deduction rule that can be ``jointed'' to the root note of the tree.
A deduction rule can be jointed to the root node of a tree if one of the premises of that deduction rule can be identified with the root node.
Then, the generator updates the tree by jointing this chosen deduction rule.
The generator continues this step multiple times until the tree achieves the required depth.

The backward random deduction is done as follows.
For each step, the generator randomly chooses a leaf node of the tree.
Then, the generator randomly chooses a deduction rule that can be jointed to the leaf node.
Here, a deduction rule can be jointed to the leaf node if the deduction rule's conclusion can be identified with the leaf node.
Then, the generator updates the tree by jointing this chosen deduction rule.
The generator continues this step multiple times until the complexity of branches achieves the required level.

\subsection{Factual Distractor Generation}   \label{appendix:sec:FLD_distractor_generation}

We implemented three types of distractors.
For a deductive nstance, we use the random mixture of these distractors.
Below, we detail each type of distractor.

\noindent  \textbf{Logical Distractor:}
We construct a distractive formula the form of which is similar to a gold formula.
For example, if a gold formula is $(A \land B) \rightarrow C$, then the following formula can be a distractor: $(\neg A \land B) \rightarrow C$.
The aim of this type of distractor is to generate negative facts in a logic sense.

\noindent  \textbf{Linguistic Distractor:}
We construct a distractive sentence by randomly flipping a word in a gold sentence into another word.
For example, if a gold sentence is ``If it is not the fact that a sun rises, then (\dots)'', then  the following sentence can be a distractor: ``If it is not the lion that a sun rises, (\dots)''
We considered grammatical constraints (e.g., part-of-speech) when flipping a word.
Note that these distractors are made after the natural languages are assigned to gold formulas.
The aim of this type of distractor is to generate negative facts in a linguistic sense.

\noindent  \textbf{Negative Tree Distractor:}
We create another proof tree irrelevant to the gold proof tree and use its leaf nodes as distractors.
If a prover chooses these distractors as the starting point of the proof, then it will reach a conclusion that is irrelevant to the given hypothesis.
Thus, this type of distractor measures the prover's look-ahead ability.

\subsection{Natural Language Assignment}   \label{appendix:sec:FLD_NL_assigner}

We show an example of the natural language template file in \cref{appendix:fig:argument_template}.

When we assign natural language statements to each atomic component such as $A, B, F, G, H, I, a, b$, we used grammatical heuristics as exemplified as follows:
(i) Atomic propositions like $A$ and $B$ are converted into complete-sentence statement like ``[NOUN] is [ADJ]'', ``[NOUN] [VERB]'' and ``[NOUN] occurs''.
(ii) (Logical) predicates like $F$ and $G$ are converted into predicate phrases such as ``[VERB]'', ``is [ADJ]'' and ``is [NOUN]''.
(iii) Constants like $a$ are converted into entity-like phrases such as ``[NOUN]''.

\section{Details of Experiments} \label{appendix:sec:experiments}

\subsection{Corpora}  \label{appendix:sec:corpora}

As shown in \cref{tb:corpora}, we generated multiple corpora to align conditions as closely as possible across the corpora being compared.
Below, we detail each aspect.

\paragraph{Preprocessing on RuleTaker:}
We sub-sampled the examples so that the number of examples in the training split becomes 30k and that the answer label distribution (over ``proved'', ``disproved'', and ``unknown'') becomes uniform.

\paragraph{Dataset Size:}
All the corpora have the training split of 30k examples.
FLD corpora have validation and test split of 1k examples.
For RuleTaker, see \cite{tafjord-etal-2021-proofwriter}.

\paragraph{Label Distribution:}

All the corpora have a uniform distribution over answer labels, i.e., over ``proved'', ``disproved'', and ``unknown''.

\paragraph{Deduction Rules:}
``Implication'' deduction rules are shown in \cref{appendix:argument:implication}.
Complicated formula versions such as $F_1(a) \land F_2(a) \rightarrow G(b)$ are also used.
For ``critical thinking'', we used the ones listed in Figure 1 in \citet{betz-etal-2021-critical}.
For ``axioms'', we used the axioms of first-order predicate logic shown in \cref{appendix:argument:axioms}.

\paragraph{Linguistic Diversity:}
First, we detail the linguistic diversity of RuleTaker corpora.
RT is classified into ``less'' since it uses only few templates for each logical statement.
RT.PR is classified into ``more'' since it includes various paraphrases of logical statements collected from human annotators.
RT.BE is classified into ``more'' since it is completely human-made and includes various paraphrases.

For FLD corpora, ``less'' means that we used only one natural language template for each logical statement and limited vocabulary (sized 100) for each POS.
For ``more'', we used several (up to five) natural language templates for each logical statement, and a large vocabulary (sized 5000) for each POS.
Note that since the templates can be nested such as:
\begin{alignSmall}
    (A \land B) \rightarrow C: & \ \text{``If $(A \land B)$, then C.''}, \ \text{``$(A \land B)$, thus C.''}\nonumber \\
    (A \land B):  & \ \text{``A, and B''}, \ \text{``B and also A''}, \dots \nonumber
\end{alignSmall}
Thus, the number of resulting patterns is combinatorially large.

\paragraph{Formula Complexity:}
For FLD corpora that have formula complexity ``simple'', we assign each compound formula such as $\mathcal{F}$ and $\mathcal{G}$ only a single atomic component as $\mathcal{F}=A$ and $\mathcal{G}=B$.
For ``complex'', we used compound formulas randomly constructed from atomic formulas with logical operators, such as $\mathcal{F}=(A\land B)$, $\mathcal{F}=\neg(A\lor \neg B)$, in addition to the ``simple'' formulas.

RuleTaker corpora use ``complex'' formula.

\paragraph{Tree Depth:}
The tree depth of ``critical thinking'' is limited to one because the critical thinking deduction rules have high-granularity, and thus cannot be easily combined to form multistep deductions.

\paragraph{Tree Depth Distribution:}
We have two types of tree depth distribution: skewed and uniform.
The skewed distribution is biased toward lower depths.
This distribution comes from the distribution of RT(D0-D3).
The uniform distribution is uniform over the depths.

\subsection{Prover Training and Performance Measurement} \label{appendix:experiments_training}

We detail the prover training and performance measurements on the benchmarks.
This experimental setting is basically the same as \citet{yang2022nlproofs}.
Thus, please refer to the study when necessary.

\subsubsection{The Prover Model}
We added a slight modification to the original model for simplicity as follows.
While the original model predicted an answer label (i.e., proved, disproved, or unknown) of a given example on the basis of the log-likelihood of the augmented proof sequences, we predict the answer label by forcing the prover to generate the label token (``\_\_PROVED\_\_'', ``\_\_DISPROVED\_\_'' or ``\_\_UNKNOWN\_\_'') at the end of proof sequences.

\subsubsection{Few-shot Transfer to Synthetic Deduction Corpora} \label{appendix:experiments_synthetic_transfer}

We first train a prover on the training split of a corpus in \cref{tb:corpora}.
Then, we train the prover on a training split of another corpus in few-shot, and after that, we measured its performance on the test split.
We used validation split for tuning hyperparameters.
We adopted T5-base for prover LM for computational efficiency.
\Cref{appendix:tb:hyperparameters_synthetic_transfer} shows the hyperparameters.
For the ``fully-fine-tuning'' setting used in \cref{sec:how_well,appendix:sec:synthetic_transfer_challengingness}, we trained a prover using all the dataset examples for 20k steps.

We run the experiments for one seed for computational reasons.
Training on a source corpus takes about ten hours on a single NVIDIA A100 (48GB) GPU.
Training on a target corpus takes a few hours on the same GPU.

\begin{table}[t!]
    \small
    \centering
    \tabcolsep 3pt
    \caption{
        The hyperparameters of prover training in the transfer experiments among deduction corpora.
        See \cite{yang2022nlproofs} or the code for other parameters.
        \label{appendix:tb:hyperparameters_synthetic_transfer}
    }

\begin{tabular}{@{}cll@{}}
\toprule
\multicolumn{1}{l}{}    & \multicolumn{1}{c}{Source corpus} & Target corpus \\ \midrule
transformer model       & T5-base                           & T5-base       \\
\# dataset examples    & 30000                             & 300           \\
steps                   & 20000                             & 2000          \\
learning rate           & 1e-4                              & 1e-4          \\
learning rate scheduler & AdamW                             & AdamW         \\
warmup steps            & 1000                              & 500           \\
batch size              & 64                                & 64            \\
gradient clipping       & 0.5                               & 0.5           \\ \bottomrule
\end{tabular}

\end{table}

\subsubsection{Transfer to EntailmentBank} \label{appendix:experiments_NL_transfer}

We first train a prover on the training split of a corpus from \cref{tb:corpora}.
Then, we train the prover on the training split of each EntailmentBank corpus.
We adopted T5-large for prover LM following \citet{yang2022nlproofs}.
The hyperparameters are basically the same as \citet{yang2022nlproofs}.
\Cref{appendix:tb:hyperparameters_NL_transfer} shows the hyperparameters.

For training the verifier, we used exactly the same setting as \citet{yang2022nlproofs}.

For the EB scorer, we used the same version as \citet{yang2022nlproofs}, that is, version v3 that was released on May 28, 2022 \footnote{\texttt{\url{https://github.com/allenai/entailment_bank}}}.

We run the experiments for six seeds.
Training on a source synthetic deduction corpus takes about one day on a single NVIDIA A100 (48GB) GPU.
Training on a target EB corpus takes about one day on the same GPU.

\begin{table}[t!]
    \small
    \centering
    \tabcolsep 3pt
    \caption{
        The hyperparameters of prover training in the transfer experiments from deduction corpora to EntailmentBank.
        See \cite{yang2022nlproofs} or the code for other parameters.
        \label{appendix:tb:hyperparameters_NL_transfer}
    }

\begin{tabular}{@{}cll@{}}
\toprule
                        & \multicolumn{1}{c}{\begin{tabular}[c]{@{}c@{}}Source \\ synthetic corpus\end{tabular}} & \multicolumn{1}{c}{\begin{tabular}[c]{@{}c@{}}Traget \\ EB corpus\end{tabular}} \\ \midrule
transformer model       & T5-large                                                                               & T5-large                                                                        \\
\# dataset examples    & 30000                                                                                  & 1313                                                                            \\
steps                   & 10000                                                                                  & 10000                                                                           \\
learning rate           & 1e-4                                                                                   & 5e-5(task1), 2.5e-5(task2)                                                        \\
learning rate scheduler & AdamW                                                                                  & AdamW                                                                           \\
warmup steps            & 1000                                                                                   & 1000(task1), 3000(task2)                                                        \\
batch size              & 64                                                                                     & 64                                                                              \\
gradient clipping       & 0.5                                                                                    & 0.5                                                                             \\ \bottomrule
\end{tabular}

\end{table}

\subsection{License of used Datasets.}
All the datasets used in this paper are publicly available: RuleTaker \cite{clark2020transformers,tafjord-etal-2021-proofwriter}, EntailmentBank \cite{dalvi-2022} and FLD (will be publicly available).

\section{How Well do LMs Solve Logic?} \label{appendix:sec:synthetic_transfer_challengingness}

\begin{table}[H]
\centering
\scriptsize
\tabcolsep 3pt

\caption{
    Proof accuracy of a prover fully fine-tuned on each corpus using all the examples.
    \label{appendix:tb:synthetic_transfer_challengingness}
}

\begin{tabular}{lllllll}
\toprule
\multicolumn{3}{c}{Skewed depth distrib.} & \multicolumn{4}{c}{Uniform depth distrib.}\\
\cmidrule(l{\tabcolsep}r{\tabcolsep}){1-3} \cmidrule(l{\tabcolsep}r{\tabcolsep}){4-7}

                       RT &                     RT.PR &                 sFLD-impl &                FLD-impl.0 &                     FLD.2 &                     FLD.3 (\coolFont{FLD}) &                     FLD.4 (\coolFont{FLD}$\star$) \\
\midrule
 \cellcolor{blue!58} 92.4 &  \cellcolor{blue!60} 93.9 &  \cellcolor{blue!49} 82.2 &  \cellcolor{blue!42} 74.6 &  \cellcolor{blue!35} 66.8 &  \cellcolor{blue!35} 66.4 &  \cellcolor{blue!10} 37.7 \\
\bottomrule

\end{tabular}

\end{table}

We analyze the challengingness of each corpus in detail by using \cref{appendix:tb:synthetic_transfer_challengingness}.

First, we look into the results on skewed depth distribution corpora.
As seen, \corpusFLDRTDpthsRT is more challenging than RuleTaker corpora.
Since the corpus design in relevant aspects is aligned between RuleTaker and \corpusFLDRTDpthsRT as in \cref{tb:corpora}, the difference should come from the other implementation details.
For example, the distractors of FLD are designed to be easily confused with positive facts (\cref{sec:FLD_distractor_generation}), 
 and natural language assignments are extremely diverse due to the random statement generation (\cref{sec:FLD_NL_assigner}).

\corpusFLDRTZero is more challenging than \corpusFLDRTDpthsRT even though they use the same deduction rules.
This should be because \corpusFLDRTZero has ``uniform'' tree depth distribution and thus includes a higher depth tree than \corpusFLDRTDpthsRT do (\cref{appendix:sec:corpora}).

The reason that \corpusFLDTwo is more challenging than \corpusFLDRTZero should be as follows.
First, since a proof tree is constructed from the combination of deduction rules chosen at each level of the tree, the number of possible proof tree patterns can be estimated (very roughly) as $\mathcal{O}(\mathcal{A}^d)$, where $\mathcal{A}$ is the number of deduction rule choices at each level and $d$ is the depth of the proof tree.
Next, while \corpusFLDRTZero uses only a few deduction rules ($\mathcal{A} = 2$) of implication type shown in \cref{appendix:argument:implication}, \corpusFLDTwo uses various deduction rules ($\mathcal{A} \sim 10$) of the axioms shown in \cref{appendix:argument:axioms}.
Thus, \corpusFLDTwo includes exponentially more diverse patterns of proof trees than RuleTaker.
This makes \corpusFLDTwo  more challenging than \corpusFLDRTZero.

\corpusFLDThree is the linguistically diverse version of \corpusFLDTwo.
The challengingness of \corpusFLDThree remains almost the same as that of \corpusFLDTwo provably because LMs can solve the linguistic aspects such as paraphrasing, as discussed in \cref{sec:linguistic_diversity}.

\corpusFLDFour is the higher-depth ($d$ up to 8) version of \corpusFLDThreeWOS.
This corpus is the most challenging provably due to the exponentially combinatorially more diverse patterns proof trees coming from $\mathcal{O}(\mathcal{A}^d)$.

\subsection{Answer Accuracies on Transfer Experiments among Deduction Corpora}   \label{appendix:sec:answer_accuracies}

Below, we show the results of transfer among synthetic corpora measured by the other metric of answer accuracy.

\begin{table}[h!]
\centering
\scriptsize

\caption{
    \footnotesize
    Answer accuracy of a prover fully fine-tuned using all the dataset examples on each corpus.
    \label{tb:answer_accuracy_synthetic_transfer_challengingness}
}

\begin{tabular}{lllllll}
\toprule
                       RT &                     RT.PR &                  sFLD-impl &                  FLD-impl.0 &                     FLD.2 &                     FLD.3 &                     FLD.4 \\
\midrule
 \cellcolor{blue!58} 95.2 &  \cellcolor{blue!59} 95.8 &  \cellcolor{blue!60} 96.1 &  \cellcolor{blue!57} 94.9 &  \cellcolor{blue!46} 88.3 &  \cellcolor{blue!45} 87.7 &  \cellcolor{blue!10} 68.1 \\
\bottomrule
\end{tabular}

\end{table}

\begin{table}[h!]
\centering
\scriptsize
\tabcolsep 3.0pt

\caption{
    \captionFewShotAnswerAccuracy{deduction rules}
    \label{tb:answer_accuracy_synthetic_transfer_arguments}
}

\begin{tabular}{lllllll}
\toprule
{} &                        T5 &                        RT &                     RT.PR &                  sFLD-impl &                  sFLD-crit &               \textbf{sFLD} \\
\midrule
RT          &  \cellcolor{blue!40} 80.7 &  \cellcolor{blue!58} 95.2 &  \cellcolor{blue!56} 93.7 &  \cellcolor{blue!44} 83.6 &  \cellcolor{blue!43} 83.2 &  \cellcolor{blue!45} 84.8 \\
RT.PR       &  \cellcolor{blue!37} 78.4 &  \cellcolor{blue!56} 93.0 &  \cellcolor{blue!59} 95.8 &  \cellcolor{blue!42} 82.5 &  \cellcolor{blue!39} 79.7 &  \cellcolor{blue!42} 82.4 \\
RT.BE       &  \cellcolor{blue!10} 56.2 &  \cellcolor{blue!50} 88.3 &  \cellcolor{blue!50} 88.2 &  \cellcolor{blue!33} 75.2 &  \cellcolor{blue!39} 79.4 &  \cellcolor{blue!46} 85.0 \\
FLD (RT)    &  \cellcolor{blue!37} 78.3 &  \cellcolor{blue!43} 83.1 &  \cellcolor{blue!43} 83.3 &  \cellcolor{blue!60} 96.1 &  \cellcolor{blue!47} 86.0 &  \cellcolor{blue!58} 95.3 \\
FLD (AA)    &  \cellcolor{blue!45} 84.9 &  \cellcolor{blue!46} 85.5 &  \cellcolor{blue!45} 84.5 &  \cellcolor{blue!54} 91.7 &  \cellcolor{blue!58} 95.2 &  \cellcolor{blue!58} 95.1 \\
FLD (axiom) &  \cellcolor{blue!38} 79.0 &  \cellcolor{blue!35} 76.9 &  \cellcolor{blue!35} 76.9 &  \cellcolor{blue!48} 87.3 &  \cellcolor{blue!46} 85.7 &  \cellcolor{blue!55} 92.9 \\

\midrule
avg.        &  \cellcolor{white!35} 76.3 &  \cellcolor{white!48} 87.0 &  \cellcolor{white!48} 87.1 &  \cellcolor{white!47} 86.1 &  \cellcolor{white!45} 84.9 &  \cellcolor{white!51} \textbf{89.3} \\

\bottomrule
\end{tabular}

\end{table}

\begin{table}[h!]
\centering
\scriptsize

\caption{
    The depth-wise answer accuracies of the provers.
    \label{tb:answer_accuracy_synthetic_transfer_depth}
}

    \begin{subfigure}{0.49\linewidth}
        \centering
        \subcaption{Target corpus is \corpusFLDRTOneWOS.   \label{tb:answer_accuracy_synthetic_transfer_depth_RT}}
        
        \begin{tabular}{llll}
        \toprule
        
        {} & {}  & \multicolumn{2}{c}{Source corpus} \\
        \cmidrule(l{\tabcolsep}r{\tabcolsep}){3-4}

        {} &                        T5 &                  FLD-impl.0 &                  FLD-impl.1 \\
        \midrule
        0 &  \cellcolor{blue!35} 50.0 &  \cellcolor{blue!55} 91.7 &  \cellcolor{blue!53} 87.5 \\
        1 &  \cellcolor{blue!58} 96.8 &  \cellcolor{blue!59} 98.4 &  \cellcolor{blue!58} 96.8 \\
        2 &  \cellcolor{blue!54} 88.0 &  \cellcolor{blue!58} 97.0 &  \cellcolor{blue!57} 94.0 \\
        3 &  \cellcolor{blue!55} 90.0 &  \cellcolor{blue!56} 93.8 &  \cellcolor{blue!58} 96.2 \\
        4 &  \cellcolor{blue!54} 88.9 &  \cellcolor{blue!56} 93.3 &  \cellcolor{blue!55} 90.0 \\
        5 &  \cellcolor{blue!46} 72.6 &  \cellcolor{blue!55} 90.5 &  \cellcolor{blue!57} 95.2 \\
        6 &  \cellcolor{blue!49} 78.7 &  \cellcolor{blue!54} 89.3 &  \cellcolor{blue!59} 98.7 \\
        7 &  \cellcolor{blue!52} 84.0 &  \cellcolor{blue!55} 90.6 &  \cellcolor{blue!55} 91.5 \\
        8 &  \cellcolor{blue!51} 82.0 &  \cellcolor{blue!55} 90.2 &  \cellcolor{blue!55} 91.8 \\
        \midrule
        avg. &  \cellcolor{white!50} 81.2 &  \cellcolor{white!56} 92.8 &  \cellcolor{white!56} 93.5 \\
        \bottomrule

        \end{tabular}
        
    \end{subfigure}
    \hfill
    \begin{subfigure}{0.49\linewidth}
    
        \centering
        \subcaption{Target corpus is \corpusFLDFourWOS.  \label{tb:answer_accuracy_synthetic_transfer_depth_FLD}}

        \begin{tabular}{lll}
        \toprule
        
        {}  & \multicolumn{2}{c}{Source corpus} \\
        \cmidrule(l{\tabcolsep}r{\tabcolsep}){2-3}
        
                               T5 &                      FLD.3 &                      FLD.4 \\
        \midrule
 \cellcolor{blue!37} 75.0 &  \cellcolor{blue!60} 100.0 &  \cellcolor{blue!60} 100.0 \\
 \cellcolor{blue!43} 82.4 &   \cellcolor{blue!58} 98.6 &   \cellcolor{blue!56} 95.9 \\
 \cellcolor{blue!39} 77.3 &   \cellcolor{blue!50} 89.8 &   \cellcolor{blue!52} 91.5 \\
 \cellcolor{blue!40} 78.9 &   \cellcolor{blue!45} 84.2 &   \cellcolor{blue!40} 78.9 \\
 \cellcolor{blue!38} 76.7 &   \cellcolor{blue!41} 79.5 &   \cellcolor{blue!33} 71.2 \\
 \cellcolor{blue!32} 70.4 &   \cellcolor{blue!28} 65.7 &   \cellcolor{blue!24} 60.6 \\
 \cellcolor{blue!33} 70.6 &   \cellcolor{blue!26} 63.5 &   \cellcolor{blue!16} 51.8 \\
 \cellcolor{blue!39} 77.1 &   \cellcolor{blue!17} 52.9 &   \cellcolor{blue!11} 47.1 \\
 \cellcolor{blue!33} 71.1 &   \cellcolor{blue!11} 47.0 &   \cellcolor{blue!10} 45.2 \\
 \midrule
 \cellcolor{white!37} 75.5 &   \cellcolor{white!37} 75.7 &   \cellcolor{white!33} 71.4 \\

        \bottomrule
        \end{tabular}

    \end{subfigure}

\end{table}

\begin{table}[H]
\centering
\scriptsize

\caption{
    \captionFewShotAnswerAccuracy{the diversity of linguistic expressions}
    \label{tb:answer_accuracy_synthetic_transfer_NL}
}

\begin{tabular}{llllll}
\toprule
{} &                        T5 &                        RT &                     RT.PR &                     FLD.2 &                     FLD.3 \\
\midrule
RT    &  \cellcolor{blue!33} 80.7 &  \cellcolor{blue!58} 95.2 &  \cellcolor{blue!56} 93.7 &  \cellcolor{blue!41} 85.2 &  \cellcolor{blue!38} 83.9 \\
RT.PR &  \cellcolor{blue!29} 78.4 &  \cellcolor{blue!55} 93.0 &  \cellcolor{blue!60} 95.8 &  \cellcolor{blue!33} 80.8 &  \cellcolor{blue!35} 82.0 \\
FLD.2 &  \cellcolor{blue!17} 72.1 &  \cellcolor{blue!17} 72.1 &  \cellcolor{blue!16} 71.3 &  \cellcolor{blue!46} 88.3 &  \cellcolor{blue!44} 86.9 \\
FLD.3 &  \cellcolor{blue!10} 68.2 &  \cellcolor{blue!10} 68.0 &  \cellcolor{blue!10} 67.7 &  \cellcolor{blue!43} 86.7 &  \cellcolor{blue!45} 87.7 \\
\midrule
avg.  &  \cellcolor{white!22} 74.9 &  \cellcolor{white!35} 82.1 &  \cellcolor{white!35} 82.1 &  \cellcolor{white!41} 85.3 &  \cellcolor{white!40} 85.1 \\
\bottomrule
\end{tabular}

\end{table}

\begin{table}[h!]
\centering
\scriptsize

\caption{
    \captionFewShotAnswerAccuracy{the complexity of formulas}
    \label{tb:answer_accuracy_synthetic_transfer_logical_formula_complexity}
}

\begin{tabular}{llll}
\toprule
{} &                        T5 &                     FLD.1 &                     FLD.2 \\
\midrule
FLD.1 &  \cellcolor{blue!21} 77.9 &  \cellcolor{blue!60} 96.5 &  \cellcolor{blue!49} 91.6 \\
FLD.2 &  \cellcolor{blue!10} 72.1 &  \cellcolor{blue!21} 77.8 &  \cellcolor{blue!43} 88.3 \\
\bottomrule
\end{tabular}

\end{table}

\begin{table}[H]
\centering
\scriptsize

\caption{
    \captionFewShotAnswerAccuracy{the number of distractors}
    \label{tb:answer_accuracy_synthetic_transfer_distractors}
}

\begin{tabular}{llll}
\toprule
{} &                        T5 &                     FLD.0 &                     FLD.2 \\
\midrule
FLD.0 &  \cellcolor{blue!21} 77.7 &  \cellcolor{blue!60} 95.8 &  \cellcolor{blue!54} 93.4 \\
FLD.2 &  \cellcolor{blue!10} 72.1 &  \cellcolor{blue!33} 83.1 &  \cellcolor{blue!44} 88.3 \\
\bottomrule
\end{tabular}

\end{table}

\subsection{Results of Other Metrics on EntailmentBank}   \label{appendix:sec:EB_other_metrics}
We show the results of other metrics on EntailmentBank in \crefrange{appendix:tb:EB_task1}{appendix:tb:EB_task3}.

\begin{table*}[h]
  \scriptsize
  \centering

\caption{
    The results of all the metrics on EntailmentBank Task1.
    For the details of these metrics, refer to \citet{dalvi-etal-2021-explaining}.
    \label{appendix:tb:EB_task1}
}

\begin{tabular}{llllllll}
\toprule
 Task & \multicolumn{2}{c}{Leaves} & \multicolumn{2}{c}{Steps} & \multicolumn{2}{c}{Intermediates} & Overall \\
 \cmidrule(r){2-3} \cmidrule(r){4-5} \cmidrule(r){6-7}
{} & F1 & AllCorrect & F1 & AllCorrect & F1 & AllCorrect & AllCorrect  \\

\midrule
            T5 &  \cellcolor{blue!10} 98.2 &  \cellcolor{blue!34} 90.6 &  \cellcolor{blue!10} 53.8 &  \cellcolor{blue!10} 40.3 &      \cellcolor{blue!10} 72.1 &       \cellcolor{blue!10} 39.7 &  \cellcolor{blue!10} 36.8 \\
         RT.D5 &  \cellcolor{blue!10} 98.2 &  \cellcolor{blue!10} 88.6 &  \cellcolor{blue!60} 56.0 &  \cellcolor{blue!60} 42.7 &      \cellcolor{blue!36} 72.8 &       \cellcolor{blue!60} 41.6 &  \cellcolor{blue!60} 39.4 \\
        FLD.D5 &  \cellcolor{blue!60} 99.0 &  \cellcolor{blue!60} 92.7 &  \cellcolor{blue!48} 55.5 &  \cellcolor{blue!49} 42.2 &      \cellcolor{blue!60} 73.4 &       \cellcolor{blue!52} 41.3 &  \cellcolor{blue!56} 39.2 \\
\bottomrule

\end{tabular}

\vspace{-3mm}
\end{table*}

\begin{table*}[h]
  \scriptsize
  \centering

\caption{
    The results of all the metrics on EntailmentBank Task2.
    For the details of these metrics, refer to \citet{dalvi-etal-2021-explaining}.
    \label{appendix:tb:EB_task2}
}

\begin{tabular}{llllllll}
\toprule
 Task & \multicolumn{2}{c}{Leaves} & \multicolumn{2}{c}{Steps} & \multicolumn{2}{c}{Intermediates} & Overall \\
 \cmidrule(r){2-3} \cmidrule(r){4-5} \cmidrule(r){6-7}
{} & F1 & AllCorrect & F1 & AllCorrect & F1 & AllCorrect & AllCorrect  \\

 \midrule
            T5 &  \cellcolor{blue!10} 88.1 &  \cellcolor{blue!10} 52.1 &  \cellcolor{blue!60} 45.9 &  \cellcolor{blue!32} 33.2 &      \cellcolor{blue!60} 68.6 &       \cellcolor{blue!60} 36.2 &  \cellcolor{blue!10} 31.2 \\
         RT.D5 &  \cellcolor{blue!60} 88.6 &  \cellcolor{blue!43} 53.1 &  \cellcolor{blue!10} 45.6 &  \cellcolor{blue!10} 32.7 &      \cellcolor{blue!45} 68.4 &       \cellcolor{blue!10} 36.1 &  \cellcolor{blue!38} 32.0 \\
        FLD.D5 &  \cellcolor{blue!40} 88.4 &  \cellcolor{blue!60} 53.6 &  \cellcolor{blue!10} 45.6 &  \cellcolor{blue!60} 33.8 &      \cellcolor{blue!10} 67.9 &       \cellcolor{blue!10} 36.1 &  \cellcolor{blue!60} 32.6 \\
\bottomrule

\end{tabular}

\vspace{-3mm}
\end{table*}

\begin{table*}[h]
  \scriptsize
  \centering

\caption{
    The results of all the metrics on EntailmentBank Task3.
    For the details of these metrics, refer to \citet{dalvi-etal-2021-explaining}.
    \label{appendix:tb:EB_task3}
}

\begin{tabular}{llllllll}
\toprule
 Task & \multicolumn{2}{c}{Leaves} & \multicolumn{2}{c}{Steps} & \multicolumn{2}{c}{Intermediates} & Overall \\
 \cmidrule(r){2-3} \cmidrule(r){4-5} \cmidrule(r){6-7}
{} & F1 & AllCorrect & F1 & AllCorrect & F1 & AllCorrect & AllCorrect  \\

\midrule
            T5 &  \cellcolor{blue!60} 43.7 &   \cellcolor{blue!10} 8.2 &  \cellcolor{blue!10} 10.7 &  \cellcolor{blue!10} 6.5 &      \cellcolor{blue!29} 42.2 &       \cellcolor{blue!10} 16.8 &  \cellcolor{blue!10} 6.2 \\
         RT.D5 &  \cellcolor{blue!10} 43.1 &  \cellcolor{blue!60} 10.0 &  \cellcolor{blue!60} 13.1 &  \cellcolor{blue!57} 8.2 &      \cellcolor{blue!10} 41.7 &       \cellcolor{blue!17} 17.3 &  \cellcolor{blue!57} 8.2 \\
        FLD.D5 &  \cellcolor{blue!51} 43.6 &   \cellcolor{blue!51} 9.7 &  \cellcolor{blue!39} 12.1 &  \cellcolor{blue!60} 8.3 &      \cellcolor{blue!60} 43.0 &       \cellcolor{blue!60} 20.1 &  \cellcolor{blue!60} 8.3 \\
\bottomrule

\end{tabular}

\vspace{-3mm}
\end{table*}

\subsection{Case Study on EntailmentBank}   \label{appendix:sec:EB_case_study}

\Cref{appendix:tb:NL_transfer_case_study} shows some cases where the error of the baseline prover (T5) is fixed by the training on a deduction corpus (\corpusFLDDFiveWOS).

As seen from ``T5 error fixed by FLD.D5'' column, typical error of T5 is such as follows:
(i) T5 misses some of the premises required to derive the required conclusion, or, simply choose wrong premises.
(ii) T5 overclaims, that is, included in the generated conclusion such information that does not logically follow from the chosen premises.
It is also suggested that T5 does not understand the rules of logical operators such as negation $\neg$ and conjunction $\land$.

In contrast, the prover trained on \corpusFLDDFive captured the fundamentals of deduction rules better than the baseline:
(i) it chose correct premises necessary and sufficient to derive the next conclusion,
(ii) it included in a conclusion only such information that logically follows from the chosen premises, and
(iii) it correctly used the rules of logical operators.

\begin{table*}[t!]
\centering
\scriptsize
\tabcolsep 3.0pt

\caption{
    Cases where baseline (T5) error on a proof step is fixed by the training on \corpusFLDDFiveWOS.
    A proof step is composed of a set of chosen premises and a derived conclusion.
    A derived conclusion is either a sequence generated by the prover model when the step is an intermediate step of the proof or is (fixed to) the hypothesis given to the model when the step is the final step of the proof (marked as [hypothesis]).
    Thus, in the final step, the task of the prover model is just to choose the right premises that can derive the hypothesis.
   \label{appendix:tb:NL_transfer_case_study} 
}

\begin{tabular}{@{}llll@{}}
\toprule
                         & chosen premises                                                                                                                                                                      & derived conclusion                                                                                                                    & T5 error fixed by \corpusFLDDFive                                                                                                                                                                                            \\ \midrule
\textbf{\corpusFLDDFive} & \begin{tabular}[c]{@{}l@{}}1. force causes the speed of an object to increase / to decrease\\ 2. an increase is a kind of change\\ 3. a decrease is a kind of change\end{tabular}    & \begin{tabular}[c]{@{}l@{}}force causes the speed of an object to change\\ $[$hypothesis$]$\end{tabular}                              & \multirow{2}{*}{\begin{tabular}[c]{@{}l@{}}The chosen premises are not sufficient,\\ possibly happened because \red{T5 do not}\\ \red{understand the rule of conjunction ($\land$)}.\end{tabular}}                           \\ \cmidrule(r){1-3}
T5                       & \begin{tabular}[c]{@{}l@{}}1. force causes the speed of an object to increase / to decrease\\ 2. an increase is a kind of change\end{tabular}                                        & \begin{tabular}[c]{@{}l@{}}force causes the speed of an object to change\\ $[$hypothesis$]$\end{tabular}                              &                                                                                                                                                                                                                              \\ \midrule
\textbf{\corpusFLDDFive} & \begin{tabular}[c]{@{}l@{}}1. the milky way is made of stars\\ 2. light year can be used to measure the distance between stars\end{tabular}                                          & \begin{tabular}[c]{@{}l@{}}light year can be used to measure the distance\\ between the stars in milky way\end{tabular}               & \multirow{2}{*}{\begin{tabular}[c]{@{}l@{}}\red{T5 overclaimed}, that is,\\ included in the conclusion such information\\ that does not logically follow from\\ the chosen premises. (a.k.a ``hallucination'')\end{tabular}} \\ \cmidrule(r){1-3}
T5                       & 1. the milky way is made of stars                                                                                                                                                    & \begin{tabular}[c]{@{}l@{}}the distance between the stars in milky way\\ is light years\end{tabular}                                  &                                                                                                                                                                                                                              \\ \midrule
\textbf{\corpusFLDDFive} & \begin{tabular}[c]{@{}l@{}}1. fossils are formed by the remains of living things\\ 2. rock is a kind of nonliving thing\end{tabular}                                                 & rocks cannot form fossils $[$hypothesis$]$                                                                                            & \multirow{2}{*}{\begin{tabular}[c]{@{}l@{}}T5 missed a premise to choose wrong one,\\ possibly because \red{T5  do not understand}\\ \red{the semantics of negation ($\neg$)}\\ at ``\textit{non}living''.\end{tabular}}     \\ \cmidrule(r){1-3}
T5                       & \begin{tabular}[c]{@{}l@{}}1. fossils are formed by the remains of living things\\ 2. cannot means not be able to\end{tabular}                                                       & rocks cannot form fossils $[$hypothesis$]$                                                                                            &                                                                                                                                                                                                                              \\ \midrule
\textbf{\corpusFLDDFive} & \begin{tabular}[c]{@{}l@{}}1. the first quarter phase of the moon occurs after the new moon\\ 2. a different moon phase occurs approximately once per week\end{tabular}              & \begin{tabular}[c]{@{}l@{}}the first quarter phase of the moon will occur\\ one week after the new moon $[$hypothesis$]$\end{tabular} & \multirow{2}{*}{T5 chose wrong premises.}                                                                                                                                                                                    \\ \cmidrule(r){1-3}
T5                       & \begin{tabular}[c]{@{}l@{}}1. the first quarter phase of the moon occurs after the new moon\\ 2. old is the opposite of new\end{tabular}                                             & \begin{tabular}[c]{@{}l@{}}the first quarter phase of the moon will occur\\ one week after the new moon $[$hypothesis$]$\end{tabular} &                                                                                                                                                                                                                              \\ \midrule
\textbf{\corpusFLDDFive} & \begin{tabular}[c]{@{}l@{}}1. cutting down trees has a negative impact on an environment\\ 2. decreasing something negative has a positive impact on a thing\end{tabular}            & \begin{tabular}[c]{@{}l@{}}decreasing the amount of trees cut has\\ a positive impact on the environment\end{tabular}                 & \multirow{2}{*}{\begin{tabular}[c]{@{}l@{}}T5 generation was merely a repetition \\ of a premise (a.k.a ``repetition'')\end{tabular}}                                                                                        \\ \cmidrule(r){1-3}
T5                       & \begin{tabular}[c]{@{}l@{}}1. cutting down trees has a negative impact on an environment\\ 2. creating paper requires cutting down trees\end{tabular}                                & \begin{tabular}[c]{@{}l@{}}cutting down trees has\\ a negative impact on the environment\end{tabular}                                 &                                                                                                                                                                                                                              \\ \midrule
\textbf{\corpusFLDDFive} & \begin{tabular}[c]{@{}l@{}}1. the heart is situated in the chest\\ 2. chest and abdomen are different location on a body\end{tabular}                                                & \begin{tabular}[c]{@{}l@{}}the abdomen does not contain the heart\\ $[$hypothesis$]$\end{tabular}                                     & \multirow{2}{*}{T5 chose wrong premises.}                                                                                                                                                                                    \\ \cmidrule(r){1-3}
T5                       & \begin{tabular}[c]{@{}l@{}}1. the heart is situated in the chest\\ 2. if something does not contain something else\\      then that something lacks that something else\end{tabular} & \begin{tabular}[c]{@{}l@{}}the abdomen does not contain the heart\\ $[$hypothesis$]$\end{tabular}                                     &                                                                                                                                                                                                                              \\ \bottomrule
\end{tabular}

\end{table*}

\section{About the Version 2 of FLD}   \label{appendix:sec:release}

The current official release\footref{note1} of FLD corpora is version 2.
In version 2, we fixed a proof inconsistency issue of version 1; specifically, the version 1 corpus contained some examples with contradicting proofs, such as "proved" versus "disproved."
Due to this issue, the prover's performance had been underestimated.
Fixing this issue, the performance of the provers on version 2 differs from version 1, as follows:

\begin{table}[H]
\centering
\scriptsize

\caption{
    \small
    Proof accuracy of a prover fully fine-tuned using all the dataset examples on each corpus. Each corpus is version 2.
    \label{tb:synthetic_transfer_challengingness_version2}
}
\vspace{-2mm}

\begin{tabular}{llll}
\toprule

\multicolumn{2}{c}{FLD.v2} \\
\cmidrule(l{\tabcolsep}r{\tabcolsep}){1-2}

                        \coolFont{FLD} &                     \coolFont{FLD}$\star$ \\
\midrule
 \cellcolor{blue!35} 75.8 &  \cellcolor{blue!10} 44.4 \\
\bottomrule
\end{tabular}

\end{table}

\begin{table}[H]
\centering
\scriptsize

\caption{
    \small
    Answer accuracy of a prover fully fine-tuned using all the dataset examples on each corpus. Each corpus is version 2.
    \label{tb:synthetic_transfer_challengingness_version2}
}
\vspace{-2mm}

\begin{tabular}{llll}
\toprule

\multicolumn{2}{c}{FLD.v2} \\
\cmidrule(l{\tabcolsep}r{\tabcolsep}){1-2}

                        \coolFont{FLD} &                     \coolFont{FLD}$\star$ \\
\midrule
 \cellcolor{blue!35} 	91.6 &  \cellcolor{blue!10} 72.2
 \\
\bottomrule
\end{tabular}

\end{table}

\end{document}